\documentclass[preprint,12pt,authoryear]{elsarticle}

\usepackage{amssymb}
\usepackage{amsmath}
\usepackage{amssymb}

\usepackage{mathtools}
\DeclarePairedDelimiter\ceil{\lceil}{\rceil}

\usepackage{multirow, tabularx, booktabs}
\usepackage{array}
\newcolumntype{P}[1]{>{\centering\arraybackslash}p{#1}}
\newcolumntype{M}[1]{>{\centering\arraybackslash}m{#1}}

\usepackage[table,xcdraw,svgnames,dvipsnames]{xcolor}
\usepackage{natbib}  
\usepackage[colorlinks]{hyperref}
\AtBeginDocument{%
  \hypersetup{
    citecolor=[rgb]{0.1098,0.51765,0.69804},
    linkcolor=Green,   
    urlcolor=MidnightBlue}}

\usepackage{subfig}
\usepackage[flushleft]{threeparttable}
\usepackage{adjustbox}

\usepackage[hmargin=2cm, vmargin=3cm]{geometry}

\journal{ }

\begin{document}

\begin{frontmatter}



\title{Stacked Bidirectional and Unidirectional LSTM Recurrent Neural Network for Forecasting Network-wide Traffic State with Missing Values}


%

\author[label1]{Zhiyong Cui}
\address[label1]{Department of Civil and Environmental Engineering, University of Washington, Seattle, WA 98195 USA}
\ead{zhiyongc@uw.edu}

\author[label1]{Ruimin Ke}
\ead{ker27@uw.edu}

\author[label1]{Ziyuan Pu}
\ead{ziyuanpu@uw.edu}

\author[label1]{Yinhai Wang\corref{cor1}}
\cortext[cor1]{Corresponding author}
\ead{yinhai@uw.edu}

\begin{abstract}
Short-term traffic forecasting based on deep learning methods, especially recurrent neural networks (RNN), has received much attention in recent years. However, the potential of RNN-based models in traffic forecasting has not yet been fully exploited in terms of the predictive power of spatial-temporal data and the capability of handling missing data. In this paper, we focus on RNN-based models and attempt to reformulate the way to incorporate RNN and its variants into traffic prediction models. A stacked bidirectional and unidirectional LSTM network architecture (SBU-LSTM) is proposed to assist the design of neural network structures for traffic state forecasting. As a key component of the architecture, the bidirectional LSTM (BDLSM) is exploited to capture the forward and backward temporal dependencies in spatiotemporal data. To deal with missing values in spatial-temporal data, we also propose a data imputation mechanism in the LSTM structure (LSTM-I) by designing an imputation unit to infer missing values and assist traffic prediction. The bidirectional version of LSTM-I is incorporated in the SBU-LSTM architecture. Two real-world network-wide traffic state datasets are used to conduct experiments and published to facilitate further traffic prediction research. The prediction performance of multiple types of multi-layer LSTM or BDLSTM models is evaluated. Experimental results indicate that the proposed SBU-LSTM architecture, especially the two-layer BDLSTM network, can achieve superior performance for the network-wide traffic prediction in both accuracy and robustness. Further, comprehensive comparison results show that the proposed data imputation mechanism in the RNN-based models can achieve outstanding prediction performance when the model's input data contains different patterns of missing values.

\end{abstract}




\end{frontmatter}


\section{Introduction}
\label{sec1}

Short-term traffic forecasting based on data-driven models for ITS applications has great influence on the overall performance of modern transportation systems \cite{vlahogianni2014short}. In the last three decades, a large number of methods have been proposed for traffic forecasting in terms of predicting speed, volume, density and travel time. Studies in this area normally focus on the methodology components, aiming at developing different models to improve prediction accuracy, efficiency, or robustness. Previous literature \cite{vlahogianni2014short, MA2015187} indicates that the existing models can be roughly divided into two categories, i.e. classical statistical methods and computational intelligence (CI) approaches. Classical statistical models, such as Autoregressive Integrated Moving Average (ARIMA) and its variants \cite{williams2003modeling, chandra2009predictions}, have made great contributions to address the traffic prediction problem. With the ability to deal with high dimensional data and the capability of capturing non-linear relationship, CI approaches, especially novel machine learning methods, tend to outperform the statistical methods with respect to handling complex traffic forecasting problems \cite{karlaftis2011statistical}. The representative machine learning methods include support vector regression \cite{asif2013spatiotemporal}, K-nearest neighbor \cite{cai2016spatiotemporal}, etc. Besides, nonparametric approaches, such Kalman filter and its variants \cite{chien2003predicting, van2008online}, and matrix/tensor factorization methods \cite{tan2016short} are also widely used in traffic prediction problems.

Deep learning models, as a branch of machine learning models, become popular and rapidly be adopted in the traffic forecasting area. Most of the newly proposed traffic forecasting models \cite{MA2015187, duan2016travel, chen2016long, zhao2017lstm, wu2016short, song2016deeptransport, yu2017spatiotemporal} are based on recurrent neural networks (RNNs), which mainly process sequence data by maintaining a chain-like structure and internal memory with loops \cite{jozefowicz2015empirical}. To address RNN's exploding gradient problem, Long Short-Term Memory network (LSTM) \cite{hochreiter1997long} and the Gated Recurrent Unit network (GRU) \cite{cho2014learning} were designed to learn long-term dependencies of sequence data via gate and memory units. Many recent studies \cite{duan2016travel, chen2016long, zhao2017lstm} adopted the LSTM as a baseline or building blocks in their proposed models for traffic forecasting. Although RNN and its variants have been adopted as building blocks of traffic prediction models, few studies reformulated their model structure to improve traffic prediction accuracy and robustness. In this study, we focus on RNN-based models and attempt to design a better structure to solve the traffic prediction problem. Three primary limitations of existing RNN-based models for traffic forecasting can be summarized as follows: 1) Few existing models are capable of dealing with missing data. 2) Although time series of traffic states are normally processed in a chronological order to capture the forward dependencies, backward dependencies in traffic state sequences, which can be learned in a reverse-chronological order, has not been explored. 3) Few studies evaluate the trade-off between model capacity and complexity. 
\par

Firstly, missing data is a common problem in the traffic data collection process due to sensor or communication failure. Various data imputation methods for time series have been developed and applied to estimate missing data. However, solving the imputation and prediction tasks at the same time often leads to a two-step process where imputation and prediction models are separated \cite{che2018recurrent}. In this way, the missing patterns of the data cannot be effectively explored in prediction models, and thus may result in biased prediction results \cite{wells2013strategies}. In some real-time traffic forecasting scenarios, the assumption of data imputation methods may not be satisfied, and thus, missing data cannot be imputed in real-time. Further, it is usually computationally expensive for training and applying these imputation methods. Due to RNNs for times series with missing values have been explored and applied \cite{ che2018recurrent, lipton2016directly}, RNN-based models have the potential to combine imputation methods with prediction models. Considering the ability of LSTM to capture and maintain long-term dependencies, LSTM is even more suitable for time series imputation. {From another perspective}, the ability to impute missing values in time series can be regarded as a capability of processing unevenly spaced time series, which is unachievable for most of the LSTM-based traffic prediction models \cite{MA2015187, duan2016travel, chen2016long, song2016deeptransport, yu2017spatiotemporal}. Hence, exploiting the power of customized LSTM to predict traffic states with missing values, {as one of our main motivations}, is promising and attainable. In this study, we propose a customized LSTM structure with an imputation unit (LSTM-I) to fulfill this goal. 
\par

The second improvable aspect of previous work is the learning order of the traffic state time series in RNN-based models. Normally, the dataset fed to an LSTM model is chronologically arranged and the model's chain-like structure makes use of the forward dependencies. But in this process, it is possible that useful information does not efficiently pass through the chain-like gated structure. Therefore, it may be informative to consider backward dependencies into consideration by processing series data in a negative direction. Another reason for including backward dependencies in our study is the periodicity of the traffic states. Traffic conditions have strong periodicity and regularity, and even short-term periodicity can be observed \cite{jiang2004wavelet}. According to \cite{box2015time}, analyzing the periodicity of time series data from both forward and backward temporal perspectives will enhance the predictive performance. Besides, the impact of upstream and downstream traffic states with respect to a road segment in the traffic network should not be neglected. Previous studies \cite{chandra2009predictions, kamarianakis2010characterizing} found that past speed values of upstream and downstream have an influence on the future speed values of a location along a corridor. For complicated traffic networks with intersections and loops, upstream and downstream both refer to relative positions and two arbitrary locations can be upstream and downstream of each other. Upstream and downstream are defined with respect to space, while forward and backward dependencies are defined with respect to time. With the help of the forward and backward dependencies of spatial-temporal data, the learned features will be more comprehensive. Based on our review of the literature, few studies on traffic analysis utilized the backward dependency. To fill this gap, a bidirectional LSTM (BDLSTM) with the ability to deal with both forward and backward dependencies is adopted as a component of the proposed network framework in this study.
\par

The third limitation of previous work is the lack of trade-off evaluation between model capacity and complexity. Some newly proposed LSTM-based prediction models, such as \cite{MA2015187}, have only one LSTM layer to deal with time series. Existing studies \cite{lecun2015deep} have shown that deep LSTM architectures with several hidden layers can build up progressively higher levels of representations of sequence data. Although some studies \cite{chen2016long,wu2016short,yu2017deep} utilized more than one LSTM layers, the influence of the number of LSTM layers needs to be further evaluated. Furthermore, the impact of the number of other layers, the size of model weights, and the spatial dimension size of the network-wide traffic data should also be evaluated as influential factors of prediction performance.

\par
In this paper, we focus on RNN-based models and attempt to reformulate the way to incorporate RNNs into traffic prediction models, even when the input traffic data contains missing values. We propose a stacked bidirectional and unidirectional LSTM network architecture (SBU-LSTM) for network-wide traffic state prediction to address the aforementioned shortcomings. The evaluation of the prediction capability of stacked LSTM- or BDLSTM-based models has the potential to facilitate the further research on the deep learning model design for traffic prediction problems. Experiments based on two real-world datasets with different missing value patterns indicate that the proposed architecture can achieve outstanding prediction results. In summary, our contributions can be summarized as follows: 
\begin{enumerate}
\item We propose an LSTM structure with an imputation unit, i.e. LSTM-I, to infer and fill the missing values in the spatial-temporal input data and in return to help improve prediction accuracy. 
\item We propose a stacked bidirectional and unidirectional LSTM architecture. i.e. SBU-LSTM, for network-wide traffic forecasting. This stacked architecture with multiple layersis flexible. The evaluation of the prediction capability of stacked LSTM- or BDLSTM-based models has great potential to facilitate the design of neural network models for traffic prediction.
\item The trade-off between model capacity and complexity is evaluated and discussed.
\item Two real-world traffic state data is tested in this study and the LOOP-SEA dataset is published via \href{https://github.com/zhiyongc/Seattle-Loop-Data}{Github} \cite{cui2016new} and \href{https://zenodo.org/record/3258904#.XYjHmChKiUl}{Zenodo} \cite{yinhai_wang_2019_3258904}.
\end{enumerate}

\section{Literature Review}

\subsection{Deep Learning based Traffic Prediction}
\label{subsec1}

Deep learning-based models generated the state-of-the-art performance for traffic forecasting. Ever since the precursory study of utilizing NN into the traffic prediction problem was proposed \cite{hua1994apphcations}, many NN-based methods, like feed-forward NN \cite{gers1999learning}, fuzzy NN \cite{yin2002urban}, and recurrent NN (RNN) \cite{van2002freeway} are adopted for traffic forecasting problems more than ten years ago. Ma et al. \cite{MA2015187} firstly adopt the LSTM to forecast traffic speed. Several other studies utilize the LSTM to forecast travel time \cite{duan2016travel}, congestion \cite{chen2016long}, and traffic flow \cite{zhao2017lstm}. Song et al. \cite{song2016deeptransport} utilized shared hidden LSTM layers to help predict human mobility and transportation mode. Cui et al. \cite{cui2018deep} adopted bidirectional LSTM in the traffic prediction problem. Yu et al. \cite{yu2017spatiotemporal} combine the convolutional neural network with RNN to predict transportation states. There are many studies forecasting traffic from other perspectives, like learning traffic as images \cite{ma2017learning} or graphs \cite{cui2019traffic} by combining convolutional based neural networks.  

\subsection{Combining Imputation and Prediction}
\label{subsec2}

One option to deal with missing values is the skipping mechanism, which is usually used in the dropout process of RNNs \cite{gal2016theoretically}. The other one is data imputation. Interpolation \cite{kreindler2012effects}, and spline \cite{de1978practical} methods are simple and efficient for data imputation, but they cannot capture variable correlations and complex patterns to perform imputation. Various data imputation methods for time series, including regression \cite{chen2003detecting}, spectral analysis \cite{mondal2010wavelet}, EM algorithm \cite{garcia2010pattern}, and matrix factorization \cite{koren2009matrix} \cite{chen2019bayesian} have been developed and applied to estimate missing data. Among these non-deep learning-based models, matrix factorization methods normally can achieve state-of-the-art prediction accuracy. A new Bayesian temporal matrix factorization method \cite{sun2019bayesian} is proposed to solve spatiotemporal data prediction problems when there is missing values in the input data. This method can deal with missing values and achieve good prediction accuracy. However, the mechanisms of these methods and the sizes of the dataset used to train models are greatly different from those of deep learning-based models. Besides, combining these data imputation models with traffic prediction models often leads to a two-step process. To overcome this weakness, Che et al. \cite{che2018recurrent} firstly exploit to use a GRU-based model, GRU-D, to combine imputation and prediction models by designing a decay mechanism for data imputation.

\section{Methodology}

\subsection{Notations}

A time series of network-wide traffic states with $D$ sensor stations can be denoted as $X = \{x_1, x_2, ..., x_T\}^T\in \mathbb{R}^{T\times D}$ with $T$ time steps. Each vector $x_t \in \mathbb{R}^D$ denotes the $D$ sensors' traffic states at time $t$, whose elements $x_t^d$ represents the traffic state of $d$-th sensor station. It should be noted that the traffic state can refer to traffic speed, travel time, traffic volume, etc. In this paper,the traffic state specifically refers to traffic speed, which is consistent with the tested datasets in the experiement section.
\par
In reality, traffic sensors, like inductive loop detectors, may fail due to the breakdown of wire insulation or damage caused by construction activities or electronics unit failure. The sensor failure further will lead to missing values in the collected data. To deal with missing values, a $masking$ vector $m_t \in \{0,1\}^D$ is adopted to denote whether traffic states are missing at time step $t$. The masking vector for $x_t$ is defined as 
\begin{equation}
m_t^d=\left\{
                \begin{array}{ll}
                  1, & \text{if $x_t^d$ is observed}\\
                  0, & \text{otherwise}
                \end{array}
              \right.
\end{equation} 
Accordingly, for a traffic state data sample $X \in \mathbb{R}^{T\times D}$, we can get a masking data sample, $M = \{m_1,m_2,...,m_T\}^T \in \mathbb{R}^{T\times D}$.

\par
In this study, the traffic state prediction problem aims to learn a function $F(\cdot)$ to map $T$ steps of historical traffic state data to the next subsequent step of traffic state data, which can be described as:
\begin{equation}\label{eq:problem_state}
F([x_1, x_2,..., x_T]; [m_1,m_2,...,m_T]) = [x_{T+1}]
\end{equation}

\subsection{Long Short-Term Memory}

It has been showed that LSTMs work well on sequence-based tasks with long-term dependencies \cite{duan2016travel, chen2016long, zhao2017lstm, wu2016short, song2016deeptransport, yu2017spatiotemporal}. Although a variety of LSTM variants were proposed in recent years, a comprehensive analysis of LSTM variants shows that none of the variants can improve upon the standard LSTM architecture significantly \cite{greff2017lstm}. Thus, we adopt the LSTM as the base model in this study.
\par
\begin{figure*}[t]
\centering
\includegraphics[width=\textwidth]{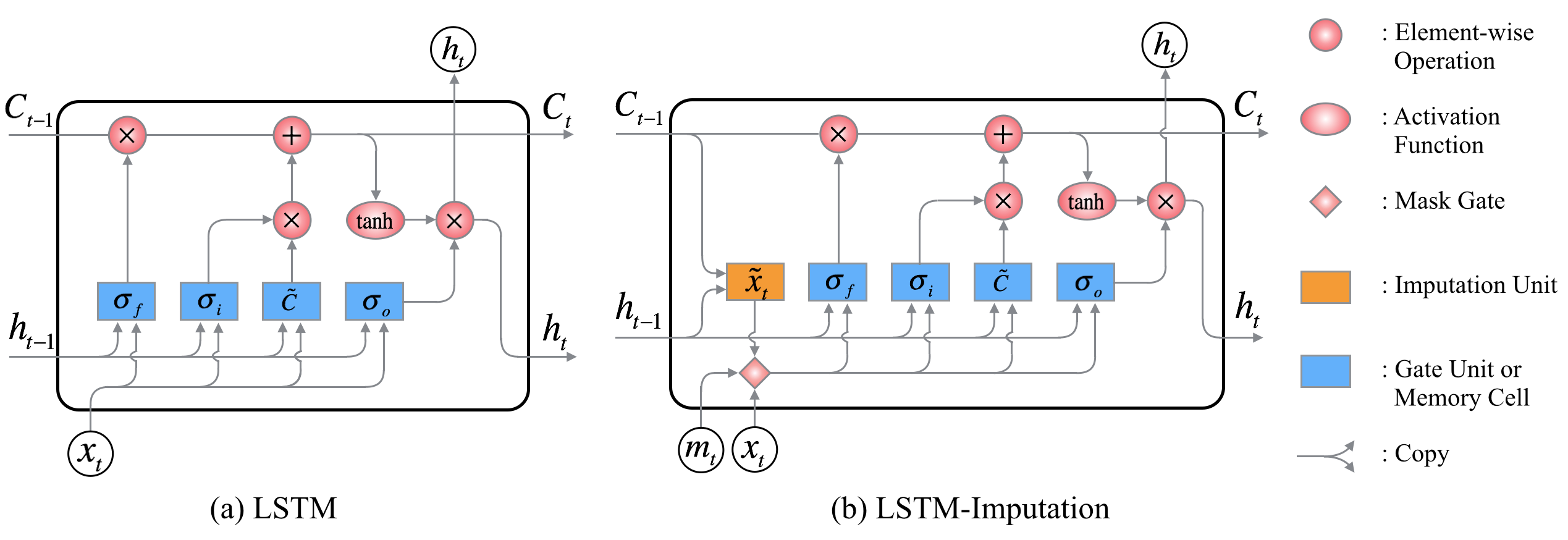}
\caption{(a) Structure of the vanilla LSTM. (b) Structure of the LSTM-I. The mask gate determines the positions of the missing values. The missing input values can be imputed via the imputation unit and the inferred/imputed values can assist the training process by adding a regularization term to the loss function.}
\label{fig:LSTM_LSTM_mask}
\end{figure*}

The structure of LSTM is shown in Figure~\ref{fig:LSTM_LSTM_mask} (a).  At time step $t$, the LSTM layer maintains a hidden memory cell $\tilde{C}_t$ and three gate units, which are input gate $i_t$, forget gate $f_t$, and output gate $o_t$. The LSTM cell takes the current variable vector $x_t$, the preceding output $h_{t-1}$, and the preceding cell state $C_{t-1}$ as inputs. With the memory cell and gate units, LSTM can learn long-term dependencies to allow useful information to pass along the LSTM network. Gate structures, especially the forget gate, help LSTM to be an effective and scalable model for sequential data learning problems \cite{greff2017lstm}. The input gate, forget gate, output gate, and memory cell in an LSTM cell are represented by blue boxes in Figure~\ref{fig:LSTM_LSTM_mask} (a). They can be calculated using the following equations:
\begin{equation} \label{eq:lstm_f_gate}
f_t = \sigma_g(W_f\cdot x_t+ U_f\cdot h_{t-1} + b_f)
\end{equation}
\begin{equation} \label{eq:lstm_i_gate}
i_t = \sigma_g(W_i\cdot x_t+ U_i\cdot h_{t-1} + b_i)
\end{equation}
\begin{equation} \label{eq:lstm_o_gate}
o_t = \sigma_g(W_o\cdot x_t+ U_o\cdot h_{t-1} + b_o)
\end{equation}
\begin{equation} \label{eq:lstm_C_gate}
\tilde{C}_t = \tanh(W_C\cdot x_t+ U_C\cdot h_{t-1} + b_C)
\end{equation}
where $\cdot$ is the matrix multiplication operator. $W_f$, $W_i,$ $W_o$, and $W_C$ are the weight matrices mapping the hidden layer input to the three gate units and the memory cell. $U_f$, $U_i$, $U_o$, and $U_C$ are the weight matrices connecting the preceding output to the three gates and the memory cell. $b_f$, $b_i$, $b_o$, and $b_C$ are four bias vectors. $\sigma_g(\cdot)$ is the gate activation function, which is a sigmoid function here, and $\tanh(\cdot)$ is the hyperbolic tangent function. Then, the cell output state $C_t$ and the layer output $h_t$ can be calculated as follows: 
\begin{equation}
C_t = f_t \odot C_{t-1} + i_t \odot \tilde{C}_t
\end{equation}
\begin{equation}
h_t = o_t \odot \tanh(C_t)
\end{equation}
where $\odot$ is the element-wise vector/matrix multiplication operator.
\par
The output of an LSTM layer can be a set of outputs from all $T$ steps, represented by $H_T=[h_1,h_2, ...,h_T]$. Here, when taking the traffic prediction problem (Equation \ref{eq:problem_state}) as an example, only the last element of the output vector $h_T$ is what we want to predict. Hence, the predicted value for the subsequent time step $T+1$ is $\hat{x}_{T+1}=h_T$. In the training process, the model's total loss $\mathcal{L}$ at each iteration can be calculated by 
\begin{equation}\label{eq:LSTM_loss}
\mathcal{L} = \text{Loss}(\hat{x}_{T+1}-x_{T+1}) = \text{Loss}(h_T-x_{T+1})
\end{equation}
where $\text{Loss}(\cdot)$ is the loss function, which is normally a mean square error function for traffic prediction problems. 

\subsection{LSTM with Imputation Unit}

For the LSTM-based prediction problem, if the input time series contains missing/null values, the model will fail due to null values cannot be computed during the training process. If the missing values are set as some pre-defined values, like zeroes, mean of historical observations, or last observed values, these biased model inputs will result in biased parameter estimation in the training processing \cite{che2018recurrent}. Further, solving the imputation and prediction tasks at the same time often results in separated imputation and prediction models.
\par
To fulfill data imputation and traffic prediction in one model, we propose an LSTM-based model with an imputation unit, called \textbf{LSTM-I}. Unlike the GRU-D \cite{che2018recurrent} targeting on inferring missing values based on the historical mean and the last observation with a learnable decay rate, the proposed LSTM-I aims to infer missing values at current time step from preceding LSTM cell states and hidden states. The weight parameters in the imputation unit are learnable. Further, the values inferred from the imputation values can contribute in the training process. In this way, the LSTM-I can complete the data imputation and prediction tasks at the same time. Thus, it is particularly suitable for online traffic prediction problems, which may frequently encounter missing values issues. Please note that the inferred values may not be the ``actual`` missing values, since the proposed imputation unit is only designed for generating appropriate values to help the calculation process in the LSTM structure work properly and generate accurate predictions.
\par
In LSTM-I, we design an imputation unit $\sigma_p$, which is fed with the preceding cell state $C_{t-1}$ and the preceding output $h_{t-1}$, to infer the values of the subsequent observation, as shown in Figure ~\ref{fig:LSTM_LSTM_mask} (b). The inferred observation $\tilde{x}_t \in \mathbb{R}^D$ is denoted as 
\begin{equation}\label{eq:imputation_gate}
 \tilde{x}_t = \sigma_g(W_I\cdot C_{t-1}+ U_I\cdot h_{t-1} + b_I)
\end{equation}
where $W_I$ and $U_I$ are the weights and $b_I$ is the bias in the imputation unit. Then, each missing element of the input vector is updated by the inferred element
\begin{equation}\label{eq:x_update}
x_t^d \leftarrow m_t^d x_t^d + (1 - m_t^d) \tilde{x}_t^d
\end{equation}
where $\tilde{x}_t^d$ is the $d$-th element of $\tilde{x}_t$. According to Equation \ref{eq:x_update}, if $x_t^d$ is missing, $m_t^d$ is zero and $x_t^d$ is imputed by $\tilde{x}_t^d$.
\par
Besides, since each masking vector $m_t$ contains the positions/indices of missing values at time step $t$, the masking vector is also fed into the model and the LSTM-I structure can be characterized as 
\begin{equation}
x_t = m_t \odot x_t + (1 - m_t) \odot \tilde{x}_t
\end{equation}
\begin{equation}
f_t = \sigma_g(W_f\cdot x_t + U_f\cdot h_{t-1} + V_f \cdot m_t + b_f)
\end{equation}
\begin{equation}
i_t = \sigma_g(W_i\cdot x_t + U_i\cdot h_{t-1} + V_i \cdot m_t + b_i)
\end{equation}
\begin{equation}
o_t = \sigma_g(W_o\cdot x_t + U_o\cdot h_{t-1} + V_o \cdot m_t + b_o)
\end{equation}
\begin{equation}
\tilde{C}_t = \tanh(W_C\cdot x_t + U_C\cdot h_{t-1} + V_C \cdot m_t + b_C)
\end{equation}
\begin{equation}
C_t = f_t \odot C_{t-1} + i_t \odot \tilde{C}_t
\end{equation}
\begin{equation}\label{eq:LSTM-I h_t}
h_t = o_t \odot \tanh(C_t)
\end{equation}
where $V_f$, $V_i$, $V_o$, and $V_C$ are weight parameters for the masking vector $m_t$ in different gates. If there are no missing values in the input data, all elements in $m_t$ are zeros and the structure of LSTM-I is identical to that of LSTM. 
\par
Furthermore, the imputation unit can contribute to the training process. According to Equation \ref{eq:imputation_gate}, at each time step $t-1$, the imputation unit infers the  input $x_t$ and generates $\tilde{x}_t$ no matter $x_t$ contains missing values or not. When $x_t^d$ is not missing, $x_t^d$ can help LSTM-I evaluate the correctness of the inferred value $\tilde{x}_t^d$ by quantifying the difference between $x_t^d$ and $\tilde{x}_t^d$. Thus, we can add a regularization term to the model's total loss (Equation \ref{eq:LSTM_loss}) at each iteration as follows
\begin{equation}
\mathcal{L} = \text{Loss}(h_T-x_{T+1}) + \lambda \sum\limits_{t=1}^T \sum\limits_{m_t^d \neq 0} |x_t^d - \tilde{x}_t^d|
\end{equation}
where $\lambda$ is the penalty term and the regularization term $\sum\limits_{t=1}^T \sum\limits_{m_t^d \neq 0} |x_t^d - \tilde{x}_t^d|$ measures the total absolute imputation error during a training iteration. By adding the regularization term to the loss, the imputation performance can be enhanced, and it has the potential to improve the model's overall prediction accuracy.

\begin{figure}[t!]
\centering
\includegraphics[width=0.65\columnwidth]{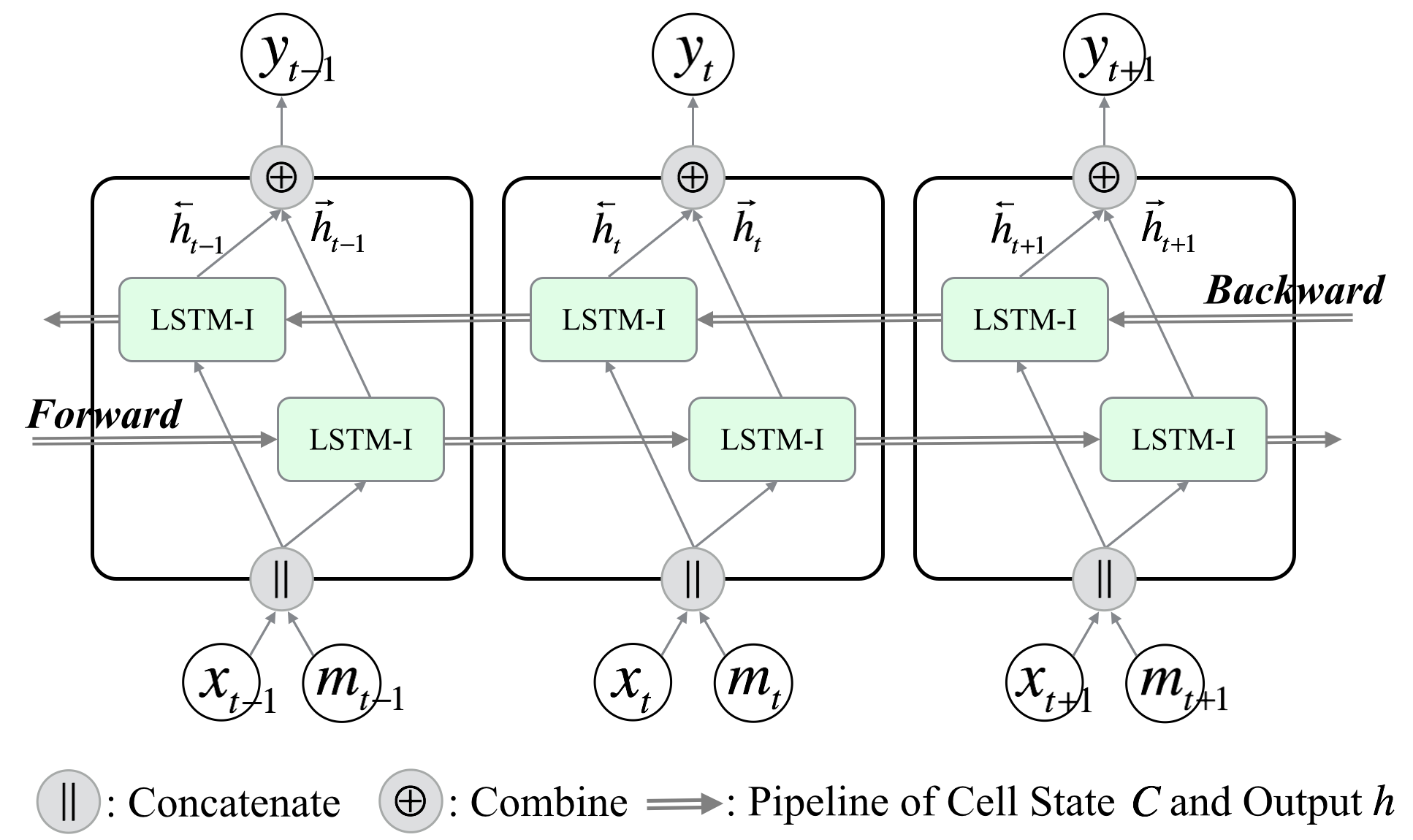}
\caption{Structure of Bi-Directional LSTM-I. }
\label{fig:BDLSTM-I}
\end{figure}

\subsection{Bidirectional LSTMs}

The idea of the BDLSTM comes from the bidirectional RNN \cite{schuster1997bidirectional}, which processes sequence data in both forward and backward directions with two separate LSTM hidden layers. It has been proved that the bidirectional networks are substantially better than unidirectional ones in many fields, like phoneme classification \cite{graves2005framewise} and speech recognition \cite{graves2013hybrid}. But, based on our review of the literature \cite{MA2015187, duan2016travel, chen2016long, wu2016short, yu2017deep}, BDLSTMs have not been utilized in traffic prediction problems. Due to the several aforementioned reasons in the introduction section, BDLSTMs are fit for handling network-wide traffic prediction problems. Thus, we adopt the BDLSTM as one component of our proposed framework.
\par
To let BDLSTMs be able to deal with time series with missing data, we propose a BDLSTM with Imputation unit, called \textbf{BDLSTM-I}, in which LSTM components are replaced with LSTM-I components, as shown in Figure~\ref{fig:BDLSTM-I}. The imputation mechanism of the LSTM-I is to infer the current missing input from the preceding cell state and output. By adopting the BDLSTM-I, missing values can be imputed from both the forward and the backward LSTM-Is. It is equivalent to infer the missing value at the current time step twice from both the preceding and the subsequent time steps, respectively. Thus, the advantage of the BDLSTM-I is that missing values are imputed based on both forward and backward temporal dependencies. Considering the periodic traffic patterns and the interaction between downstream and upstream in traffic networks, BDLSTM-I has the potentioal to reduce bias in the imputation process and further enhance traffic prediction.
\par
The structure of an unfolded BDLSTM-I layer contains a forward LSTM-I layer and a backward LSTM-I layer, which is illustrated in Figure~\ref{fig:BDLSTM-I}. The forward layer output, $\overrightarrow{h}_t$, is iteratively calculated based on positive ordered inputs $[x_1,x_2,...,x_T]$ and masks $[m_1,m_2,...,m_T]$. The backward layer output, $\overleftarrow{h}_t$, is iteratively calculated using the reversed ordered inputs and masks from time step $T$ to time step $1$. Both forward and backward outputs are calculated based on the LSTM-I model equations (Equations \ref{eq:imputation_gate} - \ref{eq:LSTM-I h_t}). The BDLSTM-I layer generates output element $y_t$ at each step $t$ based on the combination of $\overrightarrow{h}_t$ and $\overleftarrow{h}_t$ by using the following equation:
\begin{equation}
y_t = \boldsymbol{\oplus} (\overrightarrow{h}_t, \overleftarrow{h}_t)
\end{equation}
where $\boldsymbol{\oplus}$ is an average function. It should be noted that other functions, such as summation, multiply, or concatenate functions, can be used instead. Similar to the LSTM-I layer, the final output of a BDLSTM layer can be represented by a vector $Y=[y_1,y_2, ...,y_T]$. 
\par
If solely using one-layer BDLSTM-I for the prediction task, the loss function of BDLSTM-I should be defined based on that of LSTM-I. However, due to BDLSTM-I has two LSTM-I arranged in two directions, the regularization term can be slightly adjusted as 
\begin{equation}
\mathcal{L} = \text{Loss}(y_T-x_{T+1}) + \lambda \sum\limits_{t=1}^T \sum\limits_{m_t^d \neq 0} \frac{1}{2}(|x_t^d - \overrightarrow{\tilde{x}_t^d}| + |x_t^d - \overleftarrow{\tilde{x}_t^d}|)
\end{equation}
where $\overrightarrow{\tilde{x}_t^d}$ and $\overleftarrow{\tilde{x}_t^d}$ denote the inferred values from forward and backward LSTM-Is, respectively. In this way, the imputation errors of the two LSTM-Is are averaged.


\subsection{Stacked Bidirectional and Unidirectional LSTM Network Architecture}

Existing studies \cite{graves2013hybrid, lecun2015deep} have shown that LSTM architectures with several hidden layers can progressively build up a higher level of representations of sequence data, and thus, work more effectively. In a stacked multi-layer LSTM architecture, the output of a hidden layer will be fed as the input into the subsequent hidden layer. This stacking layer mechanism, which can enhance the power of neural networks, is adopted by our proposed architecture. In this study, we propose a deep architecture named stacked bidirectional and unidirectional LSTM network architecture (SBULSTM) to predict the network-wide traffic speed values. The proposed architecture does not have a fixed number of layers or use fixed types of RNNs. Instead, this architecture, possibly containing multiple layers of LSTM or BDLSTM components, can be flexible for solving different tasks. 
\par
As mentioned in previous sections, BDLSTMs can make use of both forward and backward dependencies. When feeding the spatial-temporal information of the traffic network to the BDLSTMs, both the spatial correlation of the speeds in different locations and the temporal dependencies among the traffic state sequences can be captured during the feature learning process. In this regard, a BDLSTM layer is suitable for being the first feature learning layer of a stacked architecture for network-wide traffic prediction. Meanwhile, if the input data contains missing values, a BDLSTM-I layer with the ability to deal with missing values will be used instead. 
\par

\par
For the proposed stacked architecture, the stacking/following layers could be LSTM instead of BDLSTM. Since BDLSTM contains more learnable parameters, the architecture of stacked BDLSTMs has the potential to perform better. Hence, the proposed SBU-LSTM contains a BDLSTM-I layer as the first feature-learning layer with the capability of imputing missing values. For the sake of making full use of the input data and learning complex and comprehensive features, in a SBU-LSTM architecture, the BDLSTM-I layer can be optionally stacked with one or more LSTM/BDLSTM layers. 
\par
The SBU-LSTM takes the sequence data as the input. The output $\hat{x}_{T+1}$ is generated by the last layer of SBU-LSTM. If the dataset contains missing values, the first layer of SBU-LSTM should be a BDLSTM-I. Then, the SBU-LSTM can bring the imputation errors from the BDLSTM-I layer into its loss function as a regularization term:
\begin{equation}
\mathcal{L} = \text{Loss}(\hat{x}_{T+1}-x_{T+1}) + \lambda \sum\limits_{t=1}^T \sum\limits_{m_t^d \neq 0} \frac{1}{2}(|x_t^d - \overrightarrow{\tilde{x}_t^d}| + |x_t^d - \overleftarrow{\tilde{x}_t^d}|)
\end{equation}

\section{Experiments}

\subsection{Dataset}
In this study, two real-world network-scale traffic state datasets are used for testing models. 
\subsubsection{LOOP-SEA Dataset}
The dataset named as LOOP-SEA is collected by inductive loop detectors deployed on 4 connected freeways (I-5, I-405, I-90, and SR-520) in the greater Seattle area, as shown in Figure \ref{fig:dataset_loop}. This dataset contains traffic speed data of 323 sensor stations ($D=323$). This dataset covers the whole year of 2015 and the time interval is 5 minutes. Thus, this dataset has $ 12 (5-minutes) \times 24(hour) \times 365(days)=105120$ time steps in total. For this traffic forecasting problem, if we suppose the length of the input sequence is 10 ($T$=10), the dataset contains $(105120 - 10)$ samples in total. The dataset is published on GitHub (\url{https://github.com/zhiyongc/Seattle-Loop-Data}) \cite{cui2016new} and Zenodo \cite{yinhai_wang_2019_3258904}.

\subsubsection{PEMS-BAY Dataset}

\begin{figure*}[t]
\centering
\subfloat[LOOP-SEA dataset]{
\label{fig:dataset_loop}
\includegraphics[width=0.45\textwidth, height = 0.36\textwidth ]{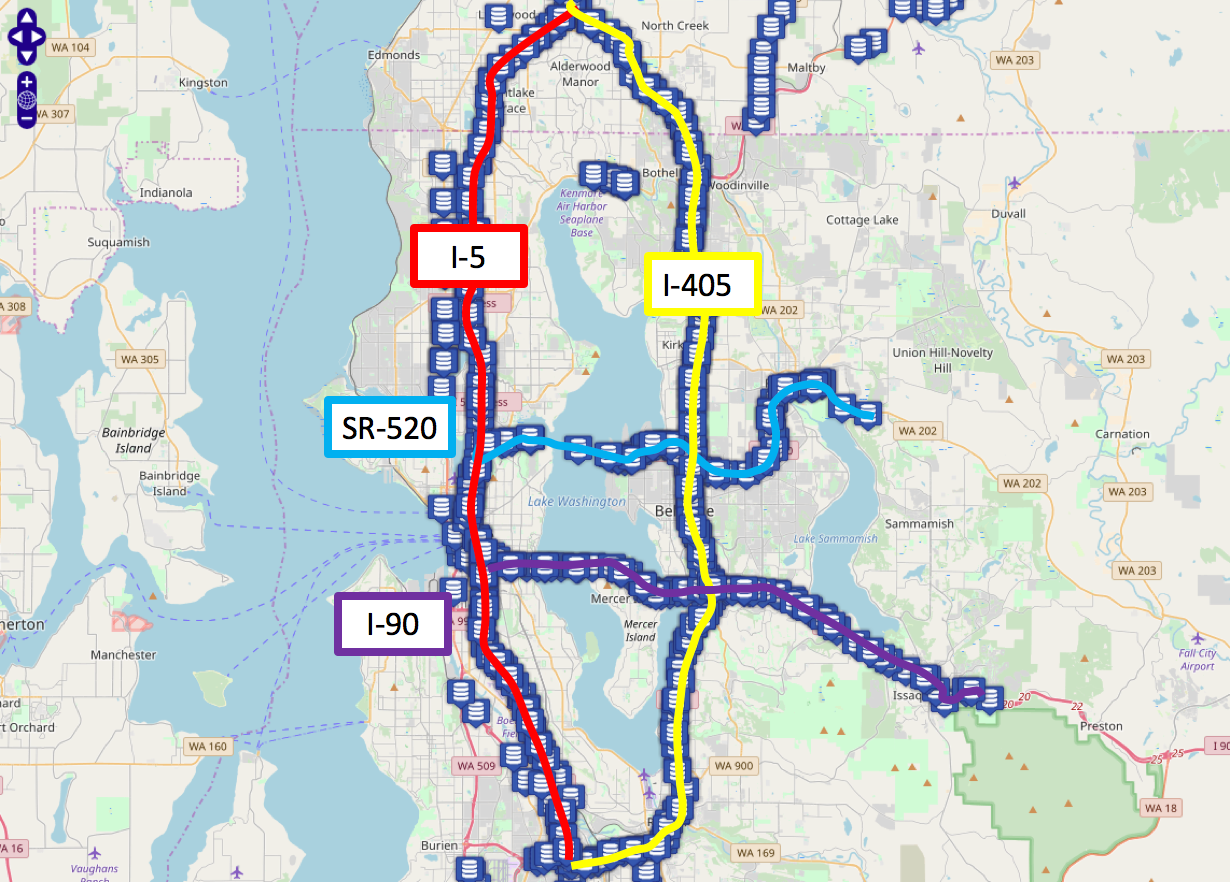}}\hfill
\subfloat[PEMS-BAY dataset]{
\label{fig:dataset_pems}
\includegraphics[width=0.45\textwidth, height = 0.36\textwidth]{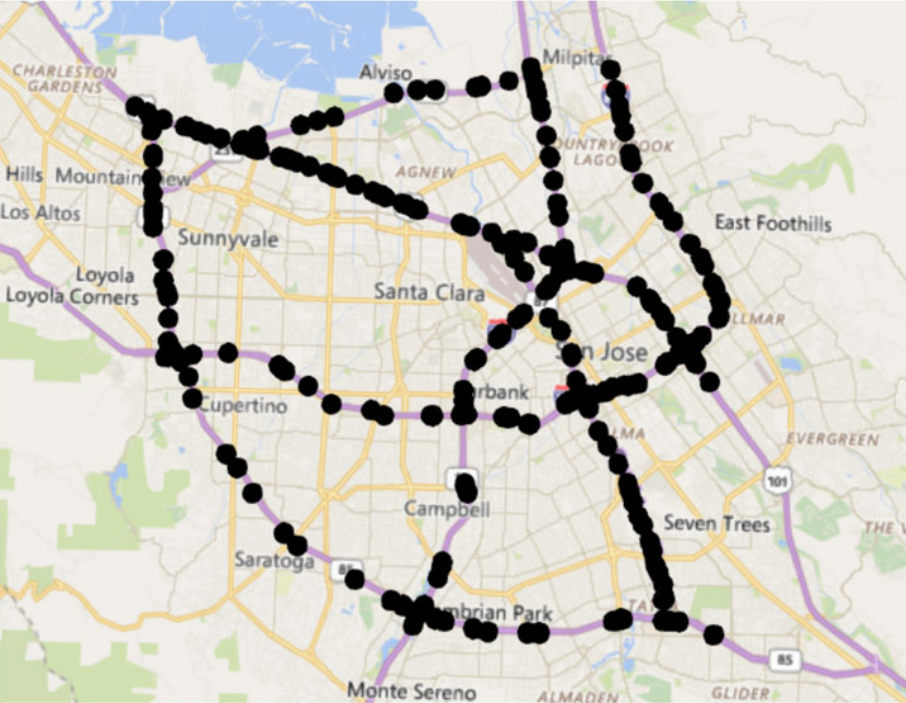}}\hfill
\caption{Datasets} 
\label{fig:datasets}
\end{figure*}

This dataset named as PEMS-BAY is collected by California Transportation Agencies (CalTrans) Performance Measurement System (PeMS). This dataset contains the speed information of 325 sensor stations in the Bay Area, as shown in Figure \ref{fig:dataset_pems}. The dataset covers six months ranging from Jan 1st, 2017 to Jun 30th, 2017. The interval of time steps is 5-minutes. The total number of observed traffic data points is 16,941,600. The dataset is published by \cite{li2018dcrnn_traffic} on the Github (\url{https://github.com/liyaguang/DCRNN}).


\subsection{Experimental Settings}

\subsubsection{Hardware}
In this study, the experiments were conducted on a computer with an Intel i7-7700 CPU @ 4.2GHz processor and 32GB of memory. All the neural network-based models are trained and evaluated on a single NVIDIA GeForece GTX 1080 Ti with 11GB Memory. 

\subsubsection{Baselines}
As indicated by multiple existing studies, the classical statistical models and machine learning models cannot outperform the LSTM model for traffic forecasting. Thus, those classical statistical models, such as ARIMA \cite{williams2003modeling} and machine learning models, such as support vector regression \cite{wu2004travel} and random forest \cite{zarei2013road}, are not compared in this study.

The compared models tested on datasets without missing values include LSTM, BDLSTM, and multiple combinations of LSTM and BDLSTM. The baseline models tested on datasets with missing values include Bayesian Gaussian CANDECOMP/PARAFAC decomposition(BGCP) \cite{chen2019bayesian}, gated recurrent unit RNN with a decay mechanism (GRU-D) \cite{che2018recurrent}, LSTM, and several combainations of LSTM-I and BDLSTM-I.


\subsubsection{Parameters}

The neural network models are implemented by PyTorch 1.0.1. In the training process, we use the mini-batch training strategy. The input of the forecasting models is a 3-D vector $\bold{X}\in\mathbb{R}^{b\times T\times D}$. The batch size $b$ is set as 64 and $D$ is the number of sensors depending on the specific dataset. The length of input sequence $T$ is set as 10, which is within a reasonable range according to \cite{yu2017deep, lv2015traffic}. The samples are randomized and divided into the training, validation, and test set with the ratio 6:2:2. All the RNN-based models are trained by minimizing the mean square error (MSE) using the Adam optimization method \cite{kingma2014adam}. The early stopping mechanism is used to avoid over-fitting. If the model improvement, i.e. the descrease of the validation loss, cannot exceed a threshold, set as 0.00001 (MSE), in 5 consecutive epochs, the training process will be terminated. We also design a learning rate decay mechanism for the training process to speed up the models' convergence. The initial learning rate of all models is set as $10^{-3}$. If the model improvement cannot surpass the threshold, the learning rate will reduce an order of magnitude until it reaches $10^{-5}$.

\subsubsection{Missing Scenarios}
When forecasting models are evaluated based on traffic state data with missing values, both the amount and the distribution of missing values will affect the prediction performance. Hence, we create a \textbf{random} scenario and a \textbf{non-random} scenario to generate datasets with different missing patterns according to \cite{chen2019missing}. The random scenario is created by randomly setting a specific proportion of values in the input as zeroes. The non-random scenario is created by randomly setting the values at a specific proportion of time steps as zeroes. The masking vectors for the two scenarios can be generated accordingly. For generating datasets with different amounts of missing values, datasets with 10\%, 20\%, 40\%, and 80\% missing values are created and tested in this study. When generating datasets with missing values, we use the identical random seed to ensure all models are evaluated on the identical datasets. 

\subsubsection{Evaluation}
To measure the effectiveness of different traffic state prediction algorithms, widely used traffic prediction metrics \cite{li2018brief}, including Mean Absolute Error (MAE), Mean Absolute Percentage Error (MAPE), and Root Mean Square Error (RMSE), are computed using the following equations:
\begin{equation}
MAE = \frac{1}{n}\sum\limits_{i=1}^n|x_i-\hat{x}_i|
\end{equation}
\begin{equation}
MAPE = \frac{100}{n}\sum\limits_{i=1}^n|\frac{x_i-\hat{x}_i}{x_i}|
\end{equation}
\begin{equation}
RMSE = \sqrt{ \frac{\sum\limits_{i=1}^n|x_i-\hat{x}_i|^2} {n} }
\end{equation}
where $x_i$ is the observed traffic speed, and $\hat{x}_i$ is the predicted speed.

\begin{table*}[htp]
\center
\caption{Performance of RNN-Based models for network-wide traffic speed prediction on LOOP-SEA dataset}
\label{table:results_loop}
\setlength\doublerulesep{0.7pt} 
\begin{threeparttable}
\centering
\begin{adjustbox}{width=1\textwidth}
\small
\begin{tabular}{lcccccccccccc}
\toprule[1pt]\midrule[0.3pt]
\multirow{3}{*}{Models}  & \multicolumn{12}{c}{Performance of Models on LOOP-SEA Dataset} 
\\
\cmidrule(l){2-13} & \multicolumn{3}{c}{N=0} & \multicolumn{3}{c}{N=1} & \multicolumn{3}{c}{N=2} & \multicolumn{3}{c}{N=3}   
\\ 
\cmidrule(l){2-4} \cmidrule(l){5-7} \cmidrule(l){8-10} \cmidrule(l){11-13} 
& MAE & MAPE & RMSE  & MAE & MAPE & RMSE & MAE & MAPE & RMSE & MAE & MAPE & RMSE  \\ \midrule[0.5pt]
\textbf{N+1} LSTM 	            
& 3.769 & 11.021 & 6.106    & 2.389 & 5.681 & 3.562     & 2.417 & 5.800 & 3.634     & 2.643 & 6.606 & 4.052\\ \hline
\textbf{N+1} BDLSTM 	        
& 3.027 & 6.815 & 5.265     & 2.336 & 5.475 & 3.507     & 2.405 & 3.631 & 5.723     & 2.472 & 5.947 & 3.750\\ \hline
\textbf{N} BDLSTM + LSTM	
& - & - & -                 & 2.523 & 6.153 & 3.809     & 2.464 & 5.954 & 3.707     & 2.579 & 6.344 & 3.911\\ \hline
\textbf{N} LSTM + BDLSTM	
& - & - & -                 & 2.362 & 5.552 & 3.542     & 2.448 & 5.875 & 3.707     & 2.580 & 6.317 & 3.941\\ 
\midrule[0.3pt]\bottomrule[1pt]
\end{tabular}
\end{adjustbox}
\end{threeparttable}
\end{table*}

\begin{table*}[htp]
\center
\caption{Performance of RNN-Based models for network-wide traffic speed prediction on PEMS-BAY dataset}
\label{table:results_pems}
\setlength\doublerulesep{0.7pt} 
\begin{threeparttable}
\centering
\begin{adjustbox}{width=1\textwidth}
\small
\begin{tabular}{lcccccccccccc}
\toprule[1pt]\midrule[0.3pt]
\multirow{3}{*}{Models}  & \multicolumn{12}{c}{Performance of Models on PEMS-BAY Dataset} 
\\
\cmidrule(l){2-13} & \multicolumn{3}{c}{N=0} & \multicolumn{3}{c}{N=1} & \multicolumn{3}{c}{N=2} & \multicolumn{3}{c}{N=3}   
\\ 
\cmidrule(l){2-4} \cmidrule(l){5-7} \cmidrule(l){8-10} \cmidrule(l){11-13} 
& MAE & MAPE & RMSE  & MAE & MAPE & RMSE & MAE & MAPE & RMSE & MAE & MAPE & RMSE  \\ \midrule[0.5pt]
\textbf{N+1} LSTM 	            
& 3.286 & 6.530 & 4.914     & 2.315 & 3.989 & 3.085     & 2.363 & 4.131 & 3.198     & 5.444 & 13.656 & 9.185\\ \hline
\textbf{N+1} BDLSTM 	        
& 1.659 & 3.003 & 4.295     & 1.186 & 2.251 & 1.927     & 1.337 & 2.583 & 2.252     & 1.569 & 3.142 & 2.822\\ \hline
\textbf{N} BDLSTM + LSTM
& - & - & -                 & 2.509 & 4.520 & 3.476     & 2.398 & 4.223 & 3.252     & 5.618 & 13.831 & 9.192\\ \hline
\textbf{N} LSTM + BDLSTM
& - & - & -                 & 1.333 & 2.545 & 2.161     & 1.526 & 3.023 & 2.671     & 2.532 & 4.732 & 6.223\\ 
\midrule[0.3pt]\bottomrule[1pt]
\end{tabular}
\end{adjustbox}
\end{threeparttable}
\end{table*}

 \subsection{Experimental Results}


In this section, the evaluation results of stacked and bidirectional LSTM-based models tested on the LOOP-SEA and PEMS-BAY datasets are shown in Table \ref{table:results_loop} and \ref{table:results_pems}, respectively. The "N DBLSTM + LSTM" refers to a n-layer BDLSTMs with an LSTM layer as the last layer. The "N LSTM + BDLSTM" is named in the similar way. 

From the experimental results shown in the two tables, we can observe at least three main similar patterns. Firstly, compared with multi-layer LSTMs and BDLSTMs, the one-layer models performs worst, which indicates the stacking mechanism can improve the prediction performance. Among the multi-layer models, the two-layer models outperforms the models with more layers. The prediction performance decreases along with the increase of amount of layers. The two-layer BDLSTM achieves the minimum MAEs of 2.336 and 1.186 on the LOOP-SEA and PEMS-BAY datasets, respectively. The second main finding is that the BDLSTM model with a specific number of layers performs better than the LSTM model with the same number of layers. This phenomenon is perticularly evident on the results of the PEMS-BAY data. The third finding is that the BDLSTM is more suitable than the LSTM for being the last layer of a model. Although the "N BDLSTM + LSTM" has more parameters than the "N LSTM + BDLSTM", when these models has same amount of layers, the "N LSTM + BDLSTM" acheives better prediction perfromance. 

The differences between the experimental results on two datasets are also obvious. The BDLSTM and the "N LSTM + BDLSTM" can achieve much better perforamnce than other types of models on the PEMS-BAY dataset. However, the superiority of BDLSTM-based models tested on the LOOP-SEA dataset is not as evident as that on the PEMS-BAY dataset. This phenomemon may be lead by that the traffic state sequences in the LOOP-SEA dataset contain more Irregular variations.

\begin{table}[t]
\centering
\caption{Performance comparison for the SBU-LSTM with different numbers of model parameters}
\label{table:spatial_weight}
\setlength\doublerulesep{0.5pt} 
\begin{tabular}{P{55mm}cccccc}
\toprule[1pt]\midrule[0.3pt]
\multirow{2}{*}{\shortstack{Dimensions of weight matrices \\ in a two-layer BDLSTM }} & \multicolumn{3}{c}{\textbf{LOOP-SEA} } & \multicolumn{3}{c}{\textbf{PEMS-BAY} } 
\\
\cmidrule(l){2-4} \cmidrule(l){5-7}    
& MAE & MAPE & RMSE & MAE & MAPE & RMSE\\
\hline \hline
$\ceil*{1/4 D}$		    &2.457  &5.896  &3.698  &1.438 	&2.788	&2.358\\	\hline
$\ceil*{1/2 D}$		    &2.383  &5.643  &3.578  &1.250 	&2.375	&2.028\\	\hline
$D$		                	    &2.336  &3.507  &3.507  &1.186	&2.251	&1.927\\	\hline
2 $D$		            &2.324  &5.443  &3.483  &1.124	&2.127	&1.824\\	\hline
4 $D$		            &2.324  &5.436  &3.489  &1.099	&2.076	&1.791\\	
\midrule[0.3pt]\bottomrule[1pt]
\end{tabular}
\end{table}

\begin{figure}[htp]
\centering
\subfloat[]{\label{fig:training_time}\includegraphics[width= 0.48\textwidth, height = 0.36\textwidth]{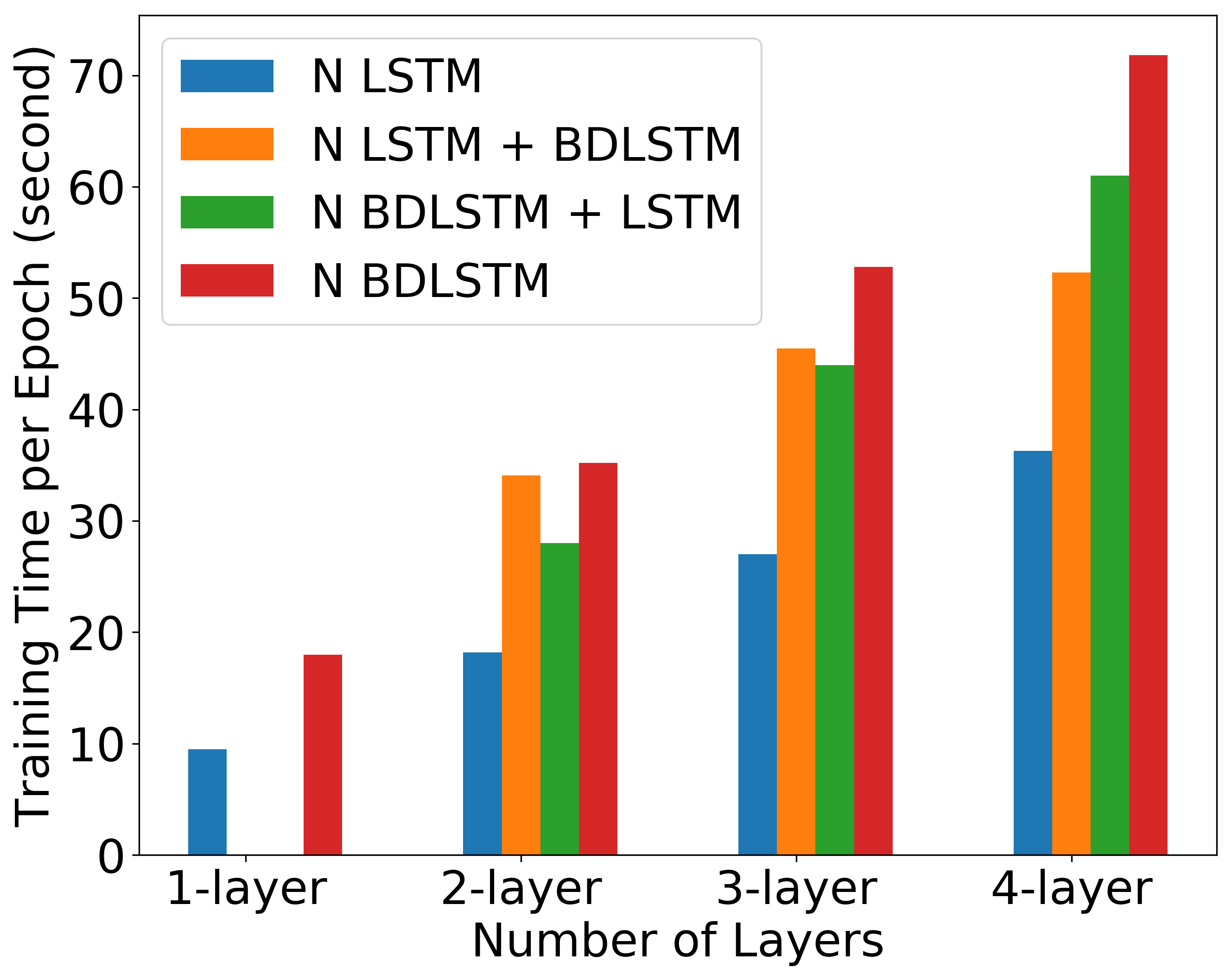}}
\centering
\subfloat[]{\label{fig:time_lags}\includegraphics[width= 0.48\textwidth, height = 0.36\textwidth]{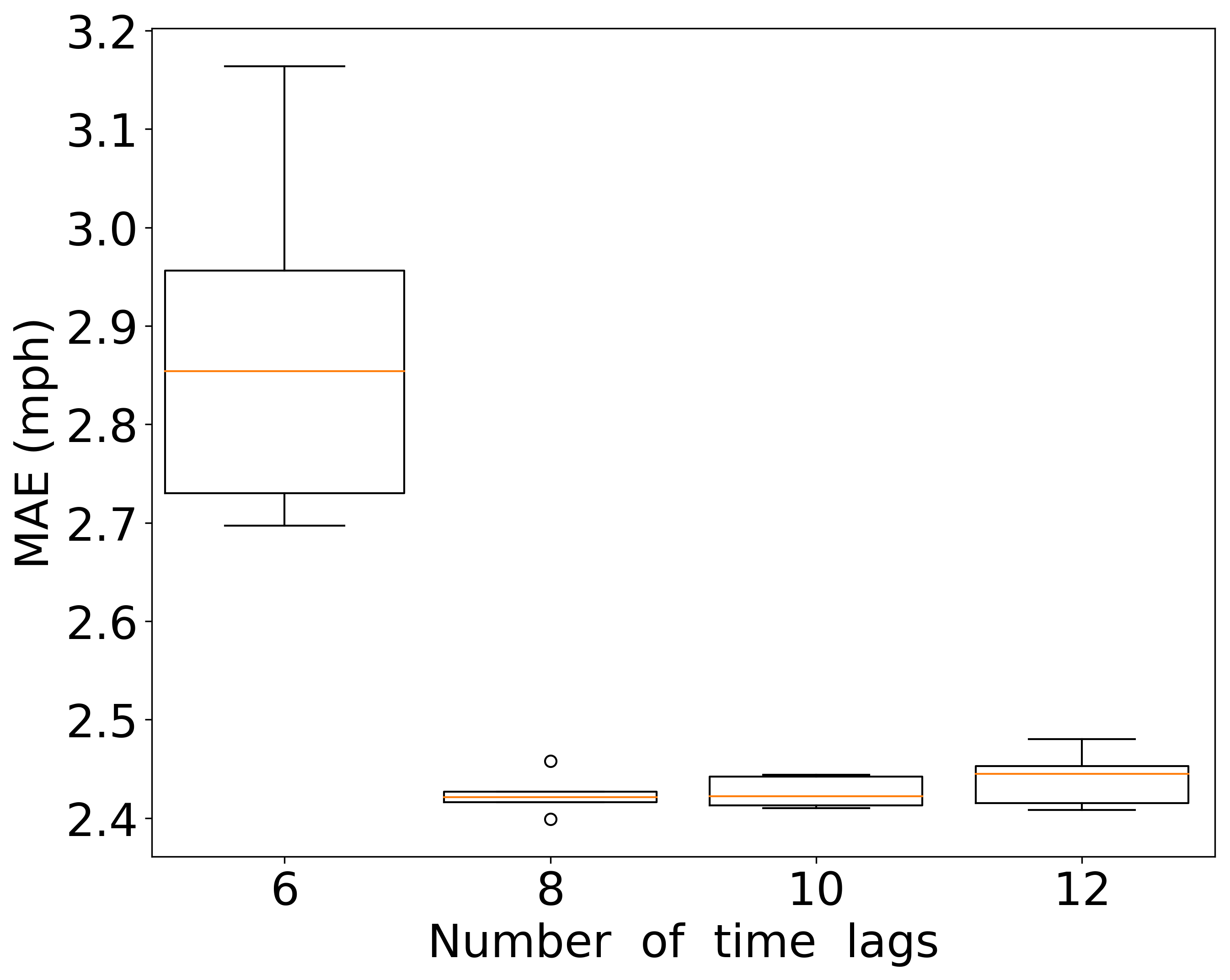}}
\caption{(a) Training time per epoch of the compared models tested on the LOOP-SEA dataset. (b) Boxplot of MAE versus number of time lags. The MAEs are generated by the BDLSTM+LSTM model tested on the LOOP-SEA dataset. The unit of one time lag (time step) is 5 minutes.}
\label{fig:5}
\end{figure}


\subsection{Training Time}
Figure \ref{fig:training_time} shows the training time per epoch of the compared models tested on the LOOP-SEA dataset. Considering the length of the input time series is fixed, the training time of a model is mostly related to the amount of parameters. Since the BDLSTM contains two LSTMs, the training time of BDLSTM is nearly double of that of LSTM. The training times of those multi-layer models are nearly linearly related to the number of layers. Since the "N BDLSTM + LSTM" and the "N LSTM + BDLSTM" are the combinations of the BDLSTM and the LSTM, their training times are between those of multi-layer BDLSTMs and LSTMs. 


\subsection{Influential Factors of the RNN-based Model}
Two factors that influene the prediction perforamnce, including the size of the model weights and the amount of time lags of the input sequences, are measured in this section.

\subsubsection{Influence of the size of model weights}
Since the spatial dimension of prediction output should be same as the spatial dimension of the input sequnces ($D$), the weight matrices in one-layer LSTM or BDLSTM, i.e. the $W$ and $U$ in Equations \ref{eq:lstm_f_gate} to \ref{eq:lstm_C_gate}, should be with the dimension of $D \times D$. For a multi-layer LSTM or BDLSTM, the dimension of the weight matrices in each layer can be customized, except for the first dimension of the weight matrices in the first layer and the second dimension of the weight matrics in the last layer that should be $D$. Hence, in this section, we change the dimensions of the customizable weight matrices in a two-layer BDLSTM to mesure the influence of the size of model weights on the prediction performance. The customized dimensions include $\ceil*{1/4 D}$, $\ceil*{1/2 D}$, $D$, $2D$, and $4D$. The experimental results generated by the two-layer BDLSTM tested on both LOOP-SEA and PEMS-BAY datasets are shown in Table \ref{table:spatial_weight}. The results indicate that the more parameters the model contains, the more accurate the predicttion results are. When the customized dimension reduces to $\frac{1}{4}D$, the prediction performance on both datasets obviously decreases. When the customized dimension increase to $4D$, the model achieves best prediciton accuracy on the PEMS-BAY dataset. However, the prediction performance cannot improve much on the LOOP-SEA dataset. This phenomenon shows that for a specific type of models, increasing the amount of parameters can improve the model's prediction capability to some extent. However, the prediction accuracy will not keep improving along with the increase of the amount of model parameters.


\subsubsection{Influence of the length of the input sequences}
The length of the input sequences has an influence on the short-term traffic forecasting performance. Figure \ref{fig:time_lags} shows the boxplot of the MAE of a BDLSTM+LSTM model versus the length of the input sequence tested on the LOOP-SEA dataset. When the length equals 8, 10, and 12, the MAEs of the predictions are very close and the deviations of these MAEs are relatively small. When the number of time lags is set as 6, the MAE is much higher, and the deviation is much larger than that of other cases. That means, given the 5-minutes time interval and the studied traffic network, 6 steps of traffic sequnce data are not long enough for modeling and predicting network-wide traffic states accurately. In summary, the number of time lags tends to influence the predictive performance, especially when the input sequences is relatively short. 



\subsection{Dealing with Missing Values}

In this section, to evaluate the effectiveness of the proposed imputation unit in the LSTM-based strucutres, we compared the prediction performance of GRU-D \cite{che2018recurrent}, LSTM-I, and BDLSTM-I. Since the previous sections show that two-layer LSTM- or BDLSTM-based two-layer models have better prediction performance, the LSTM-I and BDLSTM-I stacked with LSTM or BDLSTM is also compared. Besides, although the proposed imputation unit in LSTM-I is designed for doing data imputation tasks, we still attempt to compare the data imputation performance of the proposed models with a state-of-the-art data imputation model, i.e. the Bayesian Gaussian CANDECOMP/PARAFAC (BGCP) tensor decomposition model \cite{chen2019bayesian}. The parameters of the BGCP is identical to the parameter settings in \cite{chen2019bayesian}.

\subsubsection{Comparison with Forecasting Methods}

In this section, the compared models are tested on the LOOP-SEA and PEMS-BAY datasets with different missing scenarios, including random and non-random scenarios, and different missing rates ranging from 10\% to 80\%. The experimental results tested on the two datasets with random missing value are shown in Table \ref{table:missing_loop_random} and \ref{table:missing_pems_random}, and the results tested on datasets with the non-random missing value scenario are shown in Table \ref{table:missing_loop_nonrandom} and \ref {table:missing_pems_nonrandom}, respectively.

When the missing rate is relatively small (10\% and 20\%), the experimental results indicate that the GRU-D model cannot outperform other compared models in the random missing scenarios. The LSTM-I also cannot deal with the missing values very well, especially in the non-random missing scenarios. However, the bidirectional and stacked LSTM-models achieve better prediction accuracy on the LOOP-SEA datasets. Further, experimental results on the PEMS-BAY datasets show that the models with a BDLSTM as the last layer perform better no matter in the random or non-random missing scenario. Among the two-layer models, the prediction results of the BDLSTM-I + BDLSTM obviously outperforms those of other compared models.

When the missing rate is relatively large (40\% and 80\%), the one-layer models, including the GRU-D, LSTM-I, and BDLSTM-I, cannot compete with the two-layer models. In all cases, the two-layer models with a BDLSTM second layer perform better than those with an LSTM second layer. The BDLSTM-I + BDLSTM model achieves the smallest prediction errors. 

Overall, the prediction results in the non-random scenario are close to or slightly better than those in the random scenario. The prediction results on the LOOP-SEA dataset have larger deviations than those on the PMES-BAY dataset. The models containing a BDLSTM-I layer slightly outperform the models with an LSTM-I layer. The experimental results also indicate that the BDLSTM is appropriate to be the last layer of a model, compared with LSTM. Hence, the BDLSTM-I + BDLSTM model outperforms other models, which is particularly obvious on the PEMS-BAY dataset. 


\par 

\begin{table*}[htp]
\center
\caption{Prediction results on LOOP-SEA dataset with Random missing values}
\label{table:missing_loop_random}
\setlength\doublerulesep{0.7pt} 
\begin{threeparttable}
\centering
\begin{adjustbox}{width=1\textwidth}
\small
\begin{tabular}{lcccccccccccc}
\toprule[1pt]\midrule[0.3pt]
\multirow{3}{*}{Multi-layer Models}  & \multicolumn{12}{c}{\textbf{LOOP-SEA} dataset with \textbf{Random} missing values} 
\\
\cmidrule(l){2-13} & \multicolumn{3}{c}{Missing Rate = 10 \%} & \multicolumn{3}{c}{Missing Rate = 20 \%} & \multicolumn{3}{c}{Missing Rate = 40 \%} & \multicolumn{3}{c}{Missing Rate = 80 \%}   
\\ 
\cmidrule(l){2-4} \cmidrule(l){5-7} \cmidrule(l){8-10} \cmidrule(l){11-13} 
& MAE & MAPE & RMSE  & MAE & MAPE & RMSE & MAE & MAPE & RMSE & MAE & MAPE & RMSE  \\ \midrule[0.5pt]
GRU-D
& 3.498 & 10.354 & 5.411    & 3.676 & 10.880 & 5.881    & 3.973 & 11.760 & 6.032    & 4.539 & 13.480 & 7.292\\ \hline
LSTM-I 		        
& 3.989 & 11.830 & 6.443    & 4.060 & 12.053 & 6.548    & 4.226 & 12.715 & 6.812    & 4.791 & 14.89 & 7.697\\ \hline
LSTM-I + LSTM 
& 2.574 & 6.290 & 3.862     & 2.675 & 6.642 & 4.037     & 7.709 & 30.657 & 12.339   & 8.836 & 30.178 & 12.341\\ \hline
LSTM-I + BDLSTM	        
& 2.639 & 6.494 & 4.035     & 2.927 & 7.355 & 4.541     & 3.539 & 9.913 & 5.864     & 4.516 & 13.963 & 7.481\\ \hline
BDLSTM-I	            
& 2.605 & 6.358 & 3.887     & 2.687 & 6.642 & 4.024     & 2.846 & 7.178 & 4.307     & 6.813 & 17.31 & 13.183\\ \hline
BDLSTM-I + LSTM	        
& 2.888 & 7.471 & 4.391     & 3.010 & 7.969 & 4.618     & 3.415 & 9.632 & 5.442     & 4.546 & 13.917 & 7.368\\ \hline
BDLSTM-I + BDLSTM	        
& 2.768 & 6.920 & 4.208     & 2.843 & 7.200 & 4.384     & 3.078 & 8.149 & 4.878     & 3.675 & 10.655 & 6.073\\ 

\midrule[0.3pt]\bottomrule[1pt]
\end{tabular}
\end{adjustbox}
\end{threeparttable}
\end{table*}

\begin{table*}[htp]
\center
\caption{Prediction results on PEMS-BAY dataset with Random missing values}
\label{table:missing_pems_random}
\setlength\doublerulesep{0.7pt} 
\begin{threeparttable}
\centering
\begin{adjustbox}{width=1\textwidth}
\small
\begin{tabular}{lcccccccccccc}
\toprule[1pt]\midrule[0.3pt]
\multirow{3}{*}{Multi-layer Models}  & \multicolumn{12}{c}{\textbf{PEMS-BAY} dataset with \textbf{Random} missing values} 
\\
\cmidrule(l){2-13} & \multicolumn{3}{c}{Missing Rate = 10 \%} & \multicolumn{3}{c}{Missing Rate = 20 \%} & \multicolumn{3}{c}{Missing Rate = 40 \%} & \multicolumn{3}{c}{Missing Rate = 80 \%}   
\\ 
\cmidrule(l){2-4} \cmidrule(l){5-7} \cmidrule(l){8-10} \cmidrule(l){11-13} 
& MAE & MAPE & RMSE  & MAE & MAPE & RMSE & MAE & MAPE & RMSE & MAE & MAPE & RMSE  \\ \midrule[0.5pt]
GRU-D 
& 5.320 & 13.584 & 9.163   & 5.347 & 13.609 & 9.160   & 5.387 & 13.67 & 9.180      & 5.739 & 13.767 & 9.193\\ \hline
LSTM-I 		        
& 3.486 & 7.137 & 5.330     & 3.550 & 7.321 & 5.460     & 3.704 & 7.737 & 5.746     & 4.141 & 8.959 & 6.544\\ \hline
LSTM-I + LSTM 
& 2.511 & 4.536 & 3.435     & 2.746 & 5.178 & 3.877     & 5.628 & 14.027 & 9.303    & 5.641 & 14.056 & 9.300\\ \hline
LSTM-I + BDLSTM	        
& 1.566 & 3.054 & 2.599     & 1.621 & 3.155 & 2.653     & 2.349 & 5.122 & 4.365     & 3.111 & 7.189 & 5.943\\ \hline
BDLSTM-I	            
& 1.446 & 2.820 & 2.341     & 1.582 & 3.126 & 2.596     & 2.378 & 6.377 & 4.654     & 8.939 & 16.202 & 17.746\\ \hline
BDLSTM-I + LSTM	        
& 2.664 & 4.963 & 3.724     & 2.954 & 5.729 & 4.322     & 3.229 & 6.584 & 4.990     & 3.804 & 8.217 & 6.111\\ \hline
BDLSTM-I + BDLSTM	        
& 1.697 & 3.391 & 2.830     & 1.810 & 3.710 & 3.186     & 2.066 & 4.385 & 3.772     & 2.451 & 5.365 & 4.566\\ 

\midrule[0.3pt]\bottomrule[1pt]
\end{tabular}
\end{adjustbox}
\end{threeparttable}
\end{table*}

\begin{table*}[htp]
\center
\caption{Prediction results on LOOP-SEA dataset with non-random missing values}
\label{table:missing_loop_nonrandom}
\setlength\doublerulesep{0.7pt} 
\begin{threeparttable}
\centering
\begin{adjustbox}{width=1\textwidth}
\small
\begin{tabular}{lcccccccccccc}
\toprule[1pt]\midrule[0.3pt]
\multirow{3}{*}{Multi-layer Models}  & \multicolumn{12}{c}{\textbf{LOOP-SEA} dataset with \textbf{Non-Random} missing values} 
\\
\cmidrule(l){2-13} & \multicolumn{3}{c}{Missing Rate = 10 \%} & \multicolumn{3}{c}{Missing Rate = 20 \%} & \multicolumn{3}{c}{Missing Rate = 40 \%} & \multicolumn{3}{c}{Missing Rate = 80 \%}   
\\ 
\cmidrule(l){2-4} \cmidrule(l){5-7} \cmidrule(l){8-10} \cmidrule(l){11-13} 
& MAE & MAPE & RMSE  & MAE & MAPE & RMSE & MAE & MAPE & RMSE & MAE & MAPE & RMSE  \\ \midrule[0.5pt]
GRU-D
& 3.732 & 10.876 & 5.757    & 3.870 & 11.023 & 6.124    & 4.123 & 11.643 & 6.459    & 4.903 & 14.194 & 8.149\\ \hline
LSTM-I 		        
& 4.374 & 13.620 & 7.177     & 4.810  & 15.410 & 7.881    & 5.712 & 19.412 & 9.212    & 7.349 & 26.585 & 11.263\\ \hline
LSTM-I + LSTM 
& 2.653 & 6.643 & 4.062     & 2.788 & 7.120 & 4.310      & 3.048 & 8.143 & 4.785     & 4.327 & 13.573 & 7.112\\ \hline
LSTM-I + BDLSTM	        
& 2.665 & 6.674 & 4.041     & 2.728 & 6.921 & 4.151       & 2.898 & 7.650 & 4.467     & 3.864 & 11.829 & 6.458\\ \hline
BDLSTM-I	            
& 3.136 & 8.446 & 5.328     & 3.612 & 10.391 & 6.202     & 4.772 & 15.167 & 8.144    & 8.005 & 26.965 & 12.296\\ \hline
BDLSTM-I + LSTM	        
& 2.828 & 7.277 & 4.342     & 2.981 & 7.855 & 4.618      & 3.277 & 9.123 & 5.207      & 4.742 & 15.193 & 7.631\\ \hline
BDLSTM-I + BDLSTM	        
& 2.504 & 6.085 & 3.770     & 2.577 & 6.328 & 3.903     & 2.764 & 6.995 & 4.248     & 3.712 & 11.034 & 6.244\\ 

\midrule[0.3pt]\bottomrule[1pt]
\end{tabular}
\end{adjustbox}
\end{threeparttable}
\end{table*}

\begin{table*}[htp]
\center
\caption{Prediction results on PEMS-BAY dataset with non-random missing values}
\label{table:missing_pems_nonrandom}
\setlength\doublerulesep{0.7pt} 
\begin{threeparttable}
\centering
\begin{adjustbox}{width=1\textwidth}
\small
\begin{tabular}{lcccccccccccc}
\toprule[1pt]\midrule[0.3pt]
\multirow{3}{*}{Multi-layer Models} & \multicolumn{12}{c}{\textbf{PEMS-BAY} dataset with \textbf{Non-Random} missing values} 
\\
\cmidrule(l){2-13} & \multicolumn{3}{c}{Missing Rate = 10 \%} & \multicolumn{3}{c}{Missing Rate = 20 \%} & \multicolumn{3}{c}{Missing Rate = 40 \%} & \multicolumn{3}{c}{Missing Rate = 80 \%}   
\\ 
\cmidrule(l){2-4} \cmidrule(l){5-7} \cmidrule(l){8-10} \cmidrule(l){11-13} 
& MAE & MAPE & RMSE  & MAE & MAPE & RMSE & MAE & MAPE & RMSE & MAE & MAPE & RMSE  \\ \midrule[0.5pt]
GRU-D 
& 5.103 & 13.049 & 9.162  & 5.657 & 13.470 & 9.568   & 5.357 & 13.918 & 9.028      & 5.124 & 13.145 & 9.881\\ \hline
LSTM-I 		        
& 3.676 & 7.920 & 6.010     & 3.887 & 8.603 & 6.469     & 4.350 & 10.067 & 7.362    & 5.192 & 12.846 & 8.799\\ 
\hline
LSTM-I + LSTM 
& 2.971 & 5.783 & 4.393     & 2.847 & 5.441 & 4.137     & 2.891 & 5.569 & 4.241     & 5.261 & 12.784 & 8.677\\ 
\hline
LSTM-I + BDLSTM	        
& 1.787 & 3.703 & 3.184     & 1.921 & 4.082 & 3.504     & 2.058 & 4.520 & 3.834     & 3.097 & 7.244 & 5.698\\ 
\hline
BDLSTM-I	            
& 1.742 & 3.674 & 3.667     & 2.134 & 4.803 & 4.601     & 3.010 & 7.178 & 6.140     & 6.730 & 14.829 & 12.646\\ 
\hline
BDLSTM-I + LSTM	        
& 3.064 & 6.091 & 4.636     & 2.948 & 5.751 & 4.387     & 3.187 & 6.384 & 4.807     & 3.470 & 7.287 & 5.547\\ 
\hline
BDLSTM-I + BDLSTM	        
& 1.560 & 3.119 & 2.665     & 1.716 & 3.505 & 3.004     & 1.951 & 4.102 & 3.530     & 2.764 & 6.271 & 5.241\\ 
\midrule[0.3pt]\bottomrule[1pt]
\end{tabular}
\end{adjustbox}
\end{threeparttable}
\end{table*}

\subsubsection{Comparison with Data Imputation Methods}

As mentioned before, values generated by the proposed imputation unit ($\tilde{x}_t$) may not be the ``actual`` missing values. However, comparing data imputation performance of the proposed model with other data imputation models can still be informative. In this section, we compare the imputation performance of  BDLSTM-I+BDLSTM  and the BGCP model. The experimental results tested on both LOOP-SEA and PEMS-BAY datasets are presented in Table \ref{table:imputation}. The proposed method outperforms the BGCP when the missing rate is relatively small. The imputation performance of the BDLSTM-based model decreases along with the increase of the missing rate. However, the imputation errors of the BGCP model nearly keep the same when the missing rate varies. This is reasonable  because the data imputation mechanism of BGCP, which is a tensor decomposition-based model, greatly differs from that of deep learning models. The data imputation mechanism of a trained LSTM-I only uses a short-term sequence data as input data. However, tensor decomposition-based methods use the whole datasets to impute missing data. Besides, tensor decomposition-based models use different types of optimization methods and have more optimization iterations. In summary, although the proposed method is not designed for solving data imputation tasks and uses much fewer data, it can achieve similar imputation performance as the BGCP model. This experimental results imply that the proposed imputation unit in LSTM-I/BDLSTM-I can indirectly contribute to the traffic prediction task.

\begin{table*}[htp]
\center
\caption{Data imputation performance comparison}
\label{table:imputation}
\setlength\doublerulesep{0.7pt} 
\begin{threeparttable}
\centering
\begin{adjustbox}{width=1\textwidth}
\small
\begin{tabular}{llcccccccccccc}
\toprule[1pt]\midrule[0.3pt]
\multicolumn{2}{c}{\multirow{2}{*}{Datasets \& Models}} & \multicolumn{3}{c}{Missing Rate = 10 \%} & \multicolumn{3}{c}{Missing Rate = 20 \%} & \multicolumn{3}{c}{Missing Rate = 40 \%} & \multicolumn{3}{c}{Missing Rate = 80 \%}   
\\ 
\cmidrule(l){3-5} \cmidrule(l){6-8} \cmidrule(l){9-11} \cmidrule(l){12-14} 
 \multicolumn{2}{c}{} & MAE & MAPE & RMSE  & MAE & MAPE & RMSE & MAE & MAPE & RMSE & MAE & MAPE & RMSE  \\ \hline \hline
\multirow{2}{*}{LOOP-SEA} 
& BDLSTM-I + BDLSTM 		        
& 3.676 & 7.920 & 6.010     & 3.887 & 8.603 & 6.469     & 4.350 & 10.067 & 7.362    & 5.192 & 12.846 & 8.799\\ \cmidrule(l){2-14}
& BGCP
& 3.764 & 11.230 & 5.991     & 3.757 & 11.221 & 5.981     & 3.774 & 11.280 & 6.001     & 3.763 & 11.228 & 5.991\\ \hline \hline
\multirow{2}{*}{PEMS-BAY} 
& BDLSTM-I + BDLSTM 	
& 1.742 & 3.674 & 3.667     & 2.134 & 4.803 & 4.601     & 3.010 & 7.178 & 6.140     & 6.730 & 14.829 & 12.646\\ \cmidrule(l){2-14}
& BGCP	        
& 2.131 & 4.692 & 3.969     & 2.140 & 4.704 & 3.971     & 2.139 & 4.710 & 3.988     & 2.131 & 4.698 & 3.981\\ 
\midrule[0.3pt]\bottomrule[1pt]
\end{tabular}
\end{adjustbox}
\end{threeparttable}
\end{table*}

\begin{figure}[htp]
\centering
\subfloat[]{\label{fig:result_1}\includegraphics[width=\textwidth]{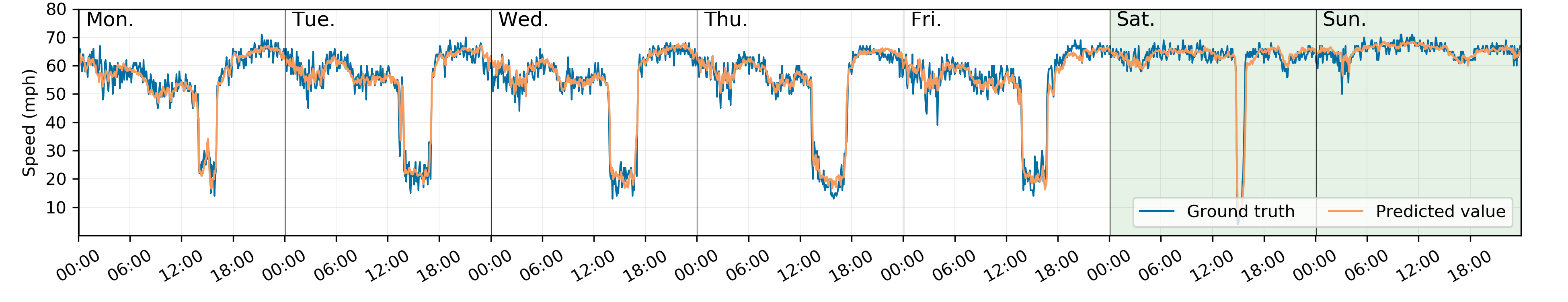}}
\\
\subfloat[]{\label{fig:result_2}\includegraphics[width=\textwidth]{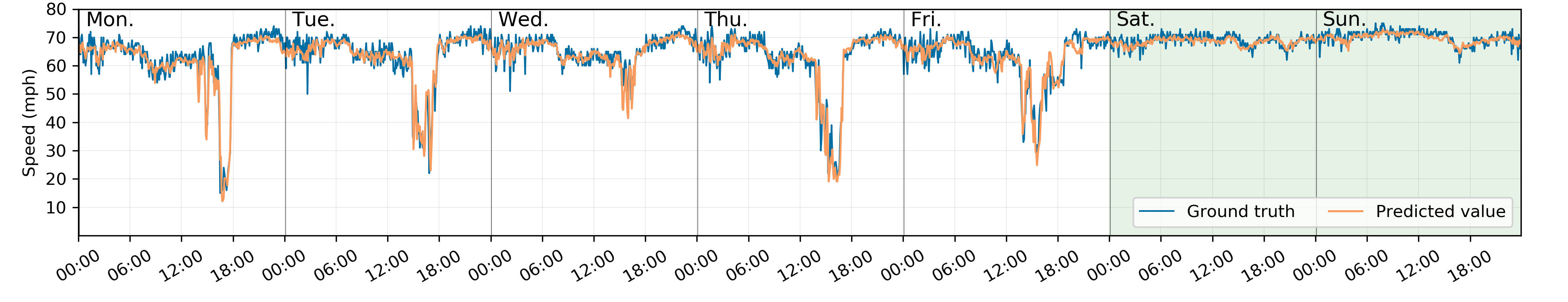}}

\subfloat[]{\label{fig:result_3}\includegraphics[width=\textwidth]{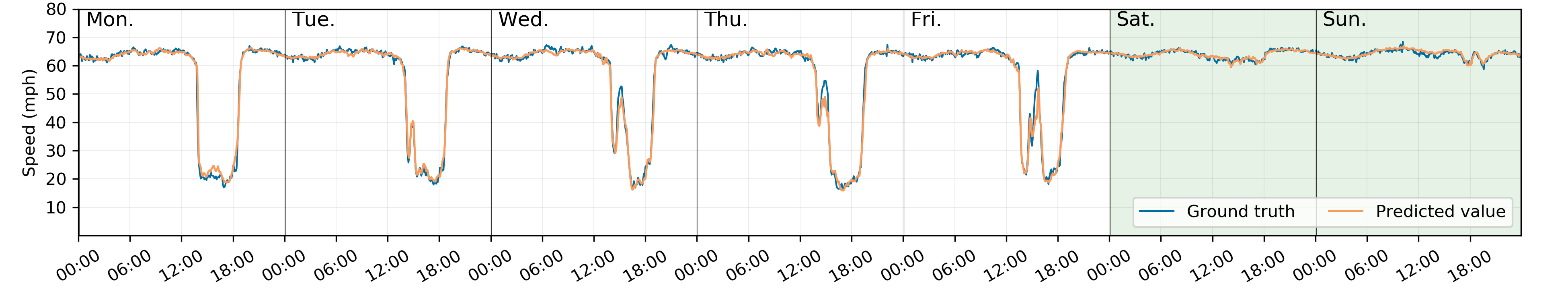}}

\subfloat[]{\label{fig:result_4}\includegraphics[width=\textwidth]{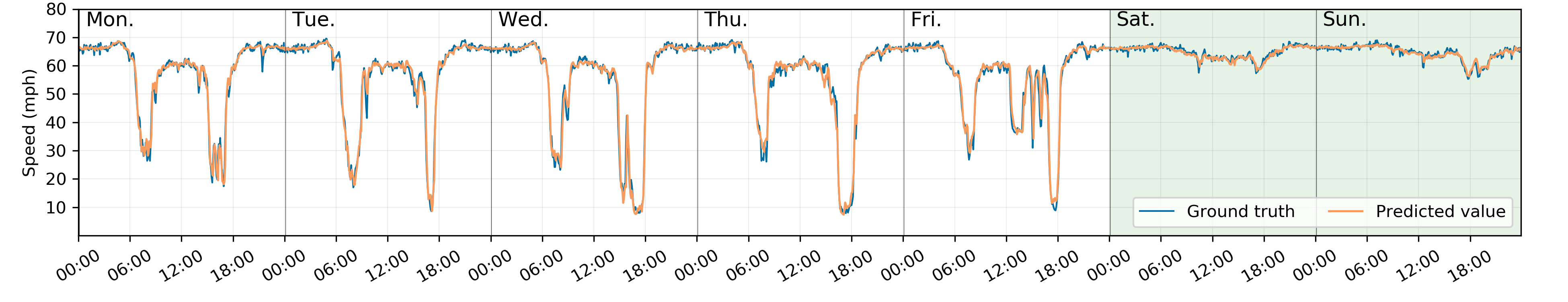}}
\caption{Visualized ground truth and predicted traffic states. Four sites are selected from the LOOP-SEA and PEMS-BAY datasets. The first two are from the LOOP-SEA dataset during the second week in 2015 and the last two are from the PEMS-BAY dataset during the fifth week in 2012. (a) Sensor ID: d005es15214. (b) Sensor ID: d005es15608. (c) Sensor ID: 400017. (d) Sensor ID: 400057.}
\label{fig:visualization}
\end{figure}

\subsection{Model Interpretation and Visualization}
In this section, we take several specific regions as examples to conduct experimental validation and visualization. Figure \ref{fig:result_1} and \ref{fig:result_2} show the ground truth and predicted traffic states of two sensing locations selected from the LOOP-SEA dataset. Figure \ref{fig:result_3} and \ref{fig:result_4} show the results of two sensing locations selected from the PEMS-BAY dataset. The traffic states in the LOOP-SEA dataset have more variations and the traffic states in the PEMS-BAY dataset varies smoothly. As shown by those figures, different traffic patterns on weekdays and weekends and the various traffic states during the peak hours can be accurately predicted.



\section{Conclusion}
In this study, we attempt to reformulate the way to incorporate LSTM into traffic prediction models. A stacked bidirectional and unidirectional LSTM  network architecture is proposed for network-wide traffic state prediction. Experimental results show that the stacked bidirectional LSTM models achieve the superior prediction performance. In addition, the evaluation of the prediction capability of multiple stacked LSTM- or BDLSTM-based models has great potential to facilitate the design of neural network models for traffic prediction problems.
\par
We also proposed an imputation unit in the LSTM model, which is designed to handle missing values. The LSTM- or BDLSTM-based models with the imputation unit can infer and fill the missing values in the spatial-temporal input data and in return to help improve prediction accuracy. Experimental results indicate that the proposed models with the imputation unit can outperform the state-of-the-art RNN based models and compete with the tensor decomposition based models. Further, the trade-off between model capacity and complexity and the influential factors of the proposed model are evaluated and discussed. In addition, two real-world traffic state data is tested in this study and the LOOP-SEA dataset is published on public accessible repositories to facilitate further research in this field. 
\par
In the future, we will carry out more in-depth studies using different datasets. Further improvements and extensions may focus on improving the model to better interpret spatial features and fusing traffic prediction with other applications. Potential applications, like non-recurring congestion detection, will be explored by combining other datasets.

\bibliographystyle{elsarticle-harv} 
\bibliography{references}

\begin{thebibliography}{56}
\providecommand{\natexlab}[1]{#1}

\bibitem[{Asif \textit{et~al.}(2013)Asif, Dauwels, Goh, Oran, Fathi, Xu,
  Dhanya, Mitrovic and Jaillet}]{asif2013spatiotemporal}
\textsc{Asif, M.~T.}, \textsc{Dauwels, J.}, \textsc{Goh, C.~Y.}, \textsc{Oran,
  A.}, \textsc{Fathi, E.}, \textsc{Xu, M.}, \textsc{Dhanya, M.~M.},
  \textsc{Mitrovic, N.} and \textsc{Jaillet, P.} (2013). Spatiotemporal
  patterns in large-scale traffic speed prediction. \textit{IEEE Transactions
  on Intelligent Transportation Systems}, \textbf{15}~(2), 794--804.

\bibitem[{Box \textit{et~al.}(2015)Box, Jenkins, Reinsel and
  Ljung}]{box2015time}
\textsc{Box, G.~E.}, \textsc{Jenkins, G.~M.}, \textsc{Reinsel, G.~C.} and
  \textsc{Ljung, G.~M.} (2015). \textit{Time series analysis: forecasting and
  control}. John Wiley \& Sons.

\bibitem[{Cai \textit{et~al.}(2016)Cai, Wang, Lu, Chen, Ding and
  Sun}]{cai2016spatiotemporal}
\textsc{Cai, P.}, \textsc{Wang, Y.}, \textsc{Lu, G.}, \textsc{Chen, P.},
  \textsc{Ding, C.} and \textsc{Sun, J.} (2016). A spatiotemporal correlative
  k-nearest neighbor model for short-term traffic multistep forecasting.
  \textit{Transportation Research Part C: Emerging Technologies}, \textbf{62},
  21--34.

\bibitem[{Chandra and Al-Deek(2009)}]{chandra2009predictions}
\textsc{Chandra, S.~R.} and \textsc{Al-Deek, H.} (2009). Predictions of freeway
  traffic speeds and volumes using vector autoregressive models.
  \textit{Journal of Intelligent Transportation Systems}, \textbf{13}~(2),
  53--72.

\bibitem[{Che \textit{et~al.}(2018)Che, Purushotham, Cho, Sontag and
  Liu}]{che2018recurrent}
\textsc{Che, Z.}, \textsc{Purushotham, S.}, \textsc{Cho, K.}, \textsc{Sontag,
  D.} and \textsc{Liu, Y.} (2018). Recurrent neural networks for multivariate
  time series with missing values. \textit{Scientific reports}, \textbf{8}~(1),
  6085.

\bibitem[{Chen \textit{et~al.}(2003)Chen, Kwon, Rice, Skabardonis and
  Varaiya}]{chen2003detecting}
\textsc{Chen, C.}, \textsc{Kwon, J.}, \textsc{Rice, J.}, \textsc{Skabardonis,
  A.} and \textsc{Varaiya, P.} (2003). Detecting errors and imputing missing
  data for single-loop surveillance systems. \textit{Transportation Research
  Record: Journal of the Transportation Research Board}, ~(1855), 160--167.

\bibitem[{Chen \textit{et~al.}(2019{\natexlab{a}})Chen, He, Chen, Lu and
  Wang}]{chen2019missing}
\textsc{Chen, X.}, \textsc{He, Z.}, \textsc{Chen, Y.}, \textsc{Lu, Y.} and
  \textsc{Wang, J.} (2019{\natexlab{a}}). Missing traffic data imputation and
  pattern discovery with a bayesian augmented tensor factorization model.
  \textit{Transportation Research Part C: Emerging Technologies}, \textbf{104},
  66--77.

\bibitem[{Chen \textit{et~al.}(2019{\natexlab{b}})Chen, He and
  Sun}]{chen2019bayesian}
\textsc{---}, \textsc{---} and \textsc{Sun, L.} (2019{\natexlab{b}}). A
  bayesian tensor decomposition approach for spatiotemporal traffic data
  imputation. \textit{Transportation research part C: emerging technologies},
  \textbf{98}, 73--84.

\bibitem[{Chen \textit{et~al.}(2016)Chen, Lv, Li and Wang}]{chen2016long}
\textsc{Chen, Y.-y.}, \textsc{Lv, Y.}, \textsc{Li, Z.} and \textsc{Wang, F.-Y.}
  (2016). Long short-term memory model for traffic congestion prediction with
  online open data. In \textit{Intelligent Transportation Systems (ITSC), 2016
  IEEE 19th International Conference on}, IEEE, pp. 132--137.

\bibitem[{Chien \textit{et~al.}(2003)Chien, Liu and
  Ozbay}]{chien2003predicting}
\textsc{Chien, S.~I.}, \textsc{Liu, X.} and \textsc{Ozbay, K.} (2003).
  Predicting travel times for the south jersey real-time motorist information
  system. \textit{Transportation Research Record}, \textbf{1855}~(1), 32--40.

\bibitem[{Cho \textit{et~al.}(2014)Cho, van Merrienboer, Gulcehre, Bahdanau,
  Bougares, Schwenk and Bengio}]{cho2014learning}
\textsc{Cho, K.}, \textsc{van Merrienboer, B.}, \textsc{Gulcehre, C.},
  \textsc{Bahdanau, D.}, \textsc{Bougares, F.}, \textsc{Schwenk, H.} and
  \textsc{Bengio, Y.} (2014). Learning phrase representations using rnn
  encoder--decoder for statistical machine translation. In \textit{Proceedings
  of the 2014 Conference on Empirical Methods in Natural Language Processing
  (EMNLP)}, pp. 1724--1734.

\bibitem[{Cui \textit{et~al.}(2019)Cui, Henrickson, Ke and
  Wang}]{cui2019traffic}
\textsc{Cui, Z.}, \textsc{Henrickson, K.}, \textsc{Ke, R.} and \textsc{Wang,
  Y.} (2019). Traffic graph convolutional recurrent neural network: A deep
  learning framework for network-scale traffic learning and forecasting.
  \textit{IEEE Transactions on Intelligent Transportation Systems}.

\bibitem[{Cui \textit{et~al.}(2018)Cui, Ke, Pu and Wang}]{cui2018deep}
\textsc{---}, \textsc{Ke, R.}, \textsc{Pu, Z.} and \textsc{Wang, Y.} (2018).
  Deep bidirectional and unidirectional lstm recurrent neural network for
  network-wide traffic speed prediction. \textit{arXiv preprint
  arXiv:1801.02143}.

\bibitem[{Cui \textit{et~al.}(2016)Cui, Zhang, Henrickson and
  Wang}]{cui2016new}
\textsc{---}, \textsc{Zhang, S.}, \textsc{Henrickson, K.~C.} and \textsc{Wang,
  Y.} (2016). New progress of drive net: An e-science transportation platform
  for data sharing, visualization, modeling, and analysis. In \textit{Smart
  Cities Conference (ISC2), 2016 IEEE International}, IEEE, pp. 1--2.

\bibitem[{De~Boor \textit{et~al.}(1978)De~Boor, De~Boor, Math{\'e}maticien,
  De~Boor and De~Boor}]{de1978practical}
\textsc{De~Boor, C.}, \textsc{De~Boor, C.}, \textsc{Math{\'e}maticien, E.-U.},
  \textsc{De~Boor, C.} and \textsc{De~Boor, C.} (1978). \textit{A practical
  guide to splines}, vol.~27. Springer-Verlag New York.

\bibitem[{Duan \textit{et~al.}(2016)Duan, Lv and Wang}]{duan2016travel}
\textsc{Duan, Y.}, \textsc{Lv, Y.} and \textsc{Wang, F.-Y.} (2016). Travel time
  prediction with lstm neural network. In \textit{Intelligent Transportation
  Systems (ITSC), 2016 IEEE 19th International Conference on}, IEEE, pp.
  1053--1058.

\bibitem[{Gal and Ghahramani(2016)}]{gal2016theoretically}
\textsc{Gal, Y.} and \textsc{Ghahramani, Z.} (2016). A theoretically grounded
  application of dropout in recurrent neural networks. In \textit{Advances in
  neural information processing systems}, pp. 1019--1027.

\bibitem[{Garc{\'\i}a-Laencina \textit{et~al.}(2010)Garc{\'\i}a-Laencina,
  Sancho-G{\'o}mez and Figueiras-Vidal}]{garcia2010pattern}
\textsc{Garc{\'\i}a-Laencina, P.~J.}, \textsc{Sancho-G{\'o}mez, J.-L.} and
  \textsc{Figueiras-Vidal, A.~R.} (2010). Pattern classification with missing
  data: a review. \textit{Neural Computing and Applications}, \textbf{19}~(2),
  263--282.

\bibitem[{Gers \textit{et~al.}(1999)Gers, Schmidhuber and
  Cummins}]{gers1999learning}
\textsc{Gers, F.~A.}, \textsc{Schmidhuber, J.} and \textsc{Cummins, F.} (1999).
  Learning to forget: Continual prediction with lstm.

\bibitem[{Graves \textit{et~al.}(2013)Graves, Jaitly and
  Mohamed}]{graves2013hybrid}
\textsc{Graves, A.}, \textsc{Jaitly, N.} and \textsc{Mohamed, A.-r.} (2013).
  Hybrid speech recognition with deep bidirectional lstm. In \textit{Automatic
  Speech Recognition and Understanding (ASRU), 2013 IEEE Workshop on}, IEEE,
  pp. 273--278.

\bibitem[{Graves and Schmidhuber(2005)}]{graves2005framewise}
\textsc{---} and \textsc{Schmidhuber, J.} (2005). Framewise phoneme
  classification with bidirectional lstm and other neural network
  architectures. \textit{Neural Networks}, \textbf{18}~(5-6), 602--610.

\bibitem[{Greff \textit{et~al.}(2017)Greff, Srivastava, Koutn{\'\i}k,
  Steunebrink and Schmidhuber}]{greff2017lstm}
\textsc{Greff, K.}, \textsc{Srivastava, R.~K.}, \textsc{Koutn{\'\i}k, J.},
  \textsc{Steunebrink, B.~R.} and \textsc{Schmidhuber, J.} (2017). Lstm: A
  search space odyssey. \textit{IEEE transactions on neural networks and
  learning systems}, \textbf{28}~(10), 2222--2232.

\bibitem[{Hochreiter and Schmidhuber(1997)}]{hochreiter1997long}
\textsc{Hochreiter, S.} and \textsc{Schmidhuber, J.} (1997). Long short-term
  memory. \textit{Neural computation}, \textbf{9}~(8), 1735--1780.

\bibitem[{Hua and Faghri(1994)}]{hua1994apphcations}
\textsc{Hua, J.} and \textsc{Faghri, A.} (1994). Applcations of artificial
  neural networks to intelligent vehicle-highway systems. \textit{RECORD},
  \textbf{1453}, 83.

\bibitem[{Jiang and Adeli(2004)}]{jiang2004wavelet}
\textsc{Jiang, X.} and \textsc{Adeli, H.} (2004). Wavelet
  packet-autocorrelation function method for traffic flow pattern analysis.
  \textit{Computer-Aided Civil and Infrastructure Engineering},
  \textbf{19}~(5), 324--337.

\bibitem[{Jozefowicz \textit{et~al.}(2015)Jozefowicz, Zaremba and
  Sutskever}]{jozefowicz2015empirical}
\textsc{Jozefowicz, R.}, \textsc{Zaremba, W.} and \textsc{Sutskever, I.}
  (2015). An empirical exploration of recurrent network architectures. In
  \textit{International Conference on Machine Learning}, pp. 2342--2350.

\bibitem[{Kamarianakis \textit{et~al.}(2010)Kamarianakis, Gao and
  Prastacos}]{kamarianakis2010characterizing}
\textsc{Kamarianakis, Y.}, \textsc{Gao, H.~O.} and \textsc{Prastacos, P.}
  (2010). Characterizing regimes in daily cycles of urban traffic using
  smooth-transition regressions. \textit{Transportation Research Part C:
  Emerging Technologies}, \textbf{18}~(5), 821--840.

\bibitem[{Karlaftis and Vlahogianni(2011)}]{karlaftis2011statistical}
\textsc{Karlaftis, M.~G.} and \textsc{Vlahogianni, E.~I.} (2011). Statistical
  methods versus neural networks in transportation research: Differences,
  similarities and some insights. \textit{Transportation Research Part C:
  Emerging Technologies}, \textbf{19}~(3), 387--399.

\bibitem[{Kingma and Ba(2014)}]{kingma2014adam}
\textsc{Kingma, D.~P.} and \textsc{Ba, J.} (2014). Adam: A method for
  stochastic optimization. \textit{arXiv preprint arXiv:1412.6980}.

\bibitem[{Koren \textit{et~al.}(2009)Koren, Bell and
  Volinsky}]{koren2009matrix}
\textsc{Koren, Y.}, \textsc{Bell, R.} and \textsc{Volinsky, C.} (2009). Matrix
  factorization techniques for recommender systems. \textit{Computer},
  \textbf{42}~(8).

\bibitem[{Kreindler and Lumsden(2012)}]{kreindler2012effects}
\textsc{Kreindler, D.~M.} and \textsc{Lumsden, C.~J.} (2012). The effects of
  the irregular sample and missing data in time series analysis.
  \textit{Nonlinear Dynamical Systems Analysis for the Behavioral Sciences
  Using Real Data}, p. 135.

\bibitem[{LeCun \textit{et~al.}(2015)LeCun, Bengio and Hinton}]{lecun2015deep}
\textsc{LeCun, Y.}, \textsc{Bengio, Y.} and \textsc{Hinton, G.} (2015). Deep
  learning. \textit{nature}, \textbf{521}~(7553), 436.

\bibitem[{Li and Shahabi(2018)}]{li2018brief}
\textsc{Li, Y.} and \textsc{Shahabi, C.} (2018). A brief overview of machine
  learning methods for short-term traffic forecasting and future directions.
  \textit{SIGSPATIAL Special}, \textbf{10}~(1), 3--9.

\bibitem[{Li \textit{et~al.}(2018)Li, Yu, Shahabi and
  Liu}]{li2018dcrnn_traffic}
\textsc{---}, \textsc{Yu, R.}, \textsc{Shahabi, C.} and \textsc{Liu, Y.}
  (2018). Diffusion convolutional recurrent neural network: Data-driven traffic
  forecasting. In \textit{International Conference on Learning Representations
  (ICLR '18)}.

\bibitem[{Lipton \textit{et~al.}(2016)Lipton, Kale and
  Wetzel}]{lipton2016directly}
\textsc{Lipton, Z.~C.}, \textsc{Kale, D.} and \textsc{Wetzel, R.} (2016).
  Directly modeling missing data in sequences with rnns: Improved
  classification of clinical time series. In \textit{Machine Learning for
  Healthcare Conference}, pp. 253--270.

\bibitem[{Lv \textit{et~al.}(2015)Lv, Duan, Kang, Li and Wang}]{lv2015traffic}
\textsc{Lv, Y.}, \textsc{Duan, Y.}, \textsc{Kang, W.}, \textsc{Li, Z.} and
  \textsc{Wang, F.-Y.} (2015). Traffic flow prediction with big data: a deep
  learning approach. \textit{IEEE Transactions on Intelligent Transportation
  Systems}, \textbf{16}~(2), 865--873.

\bibitem[{Ma \textit{et~al.}(2017)Ma, Dai, He, Ma, Wang and
  Wang}]{ma2017learning}
\textsc{Ma, X.}, \textsc{Dai, Z.}, \textsc{He, Z.}, \textsc{Ma, J.},
  \textsc{Wang, Y.} and \textsc{Wang, Y.} (2017). Learning traffic as images: a
  deep convolutional neural network for large-scale transportation network
  speed prediction. \textit{Sensors}, \textbf{17}~(4), 818.

\bibitem[{Ma \textit{et~al.}(2015)Ma, Tao, Wang, Yu and Wang}]{MA2015187}
\textsc{---}, \textsc{Tao, Z.}, \textsc{Wang, Y.}, \textsc{Yu, H.} and
  \textsc{Wang, Y.} (2015). Long short-term memory neural network for traffic
  speed prediction using remote microwave sensor data. \textit{Transportation
  Research Part C: Emerging Technologies}, \textbf{54}, 187 -- 197.

\bibitem[{Mondal and Percival(2010)}]{mondal2010wavelet}
\textsc{Mondal, D.} and \textsc{Percival, D.~B.} (2010). Wavelet variance
  analysis for gappy time series. \textit{Annals of the Institute of
  Statistical Mathematics}, \textbf{62}~(5), 943--966.

\bibitem[{Schuster and Paliwal(1997)}]{schuster1997bidirectional}
\textsc{Schuster, M.} and \textsc{Paliwal, K.~K.} (1997). Bidirectional
  recurrent neural networks. \textit{IEEE Transactions on Signal Processing},
  \textbf{45}~(11), 2673--2681.

\bibitem[{Song \textit{et~al.}(2016)Song, Kanasugi and
  Shibasaki}]{song2016deeptransport}
\textsc{Song, X.}, \textsc{Kanasugi, H.} and \textsc{Shibasaki, R.} (2016).
  Deeptransport: Prediction and simulation of human mobility and transportation
  mode at a citywide level. In \textit{IJCAI}, pp. 2618--2624.

\bibitem[{Sun and Chen(2019)}]{sun2019bayesian}
\textsc{Sun, L.} and \textsc{Chen, X.} (2019). Bayesian temporal factorization
  for multidimensional time series prediction. \textit{arXiv preprint
  arXiv:1910.06366}.

\bibitem[{Tan \textit{et~al.}(2016)Tan, Wu, Shen, Jin and Ran}]{tan2016short}
\textsc{Tan, H.}, \textsc{Wu, Y.}, \textsc{Shen, B.}, \textsc{Jin, P.~J.} and
  \textsc{Ran, B.} (2016). Short-term traffic prediction based on dynamic
  tensor completion. \textit{IEEE Transactions on Intelligent Transportation
  Systems}, \textbf{17}~(8), 2123--2133.

\bibitem[{Van~Lint(2008)}]{van2008online}
\textsc{Van~Lint, J.} (2008). Online learning solutions for freeway travel time
  prediction. \textit{IEEE Transactions on Intelligent Transportation Systems},
  \textbf{9}~(1), 38--47.

\bibitem[{Van~Lint \textit{et~al.}(2002)Van~Lint, Hoogendoorn and
  Van~Zuylen}]{van2002freeway}
\textsc{---}, \textsc{Hoogendoorn, S.} and \textsc{Van~Zuylen, H.} (2002).
  Freeway travel time prediction with state-space neural networks: modeling
  state-space dynamics with recurrent neural networks. \textit{Transportation
  Research Record: Journal of the Transportation Research Board}, ~(1811),
  30--39.

\bibitem[{Vlahogianni \textit{et~al.}(2014)Vlahogianni, Karlaftis and
  Golias}]{vlahogianni2014short}
\textsc{Vlahogianni, E.~I.}, \textsc{Karlaftis, M.~G.} and \textsc{Golias,
  J.~C.} (2014). Short-term traffic forecasting: Where we are and where we're
  going. \textit{Transportation Research Part C: Emerging Technologies},
  \textbf{43}, 3--19.

\bibitem[{Wang \textit{et~al.}(2019)Wang, Ban, Cui and
  Zhu}]{yinhai_wang_2019_3258904}
\textsc{Wang, Y.}, \textsc{Ban, X.~J.}, \textsc{Cui, Z.} and \textsc{Zhu, M.}
  (2019). {Freeway Inductive Loop Detector Dataset for Network-wide Traffic
  Speed Prediction}.

\bibitem[{Wells \textit{et~al.}(2013)Wells, Chagin, Nowacki and
  Kattan}]{wells2013strategies}
\textsc{Wells, B.~J.}, \textsc{Chagin, K.~M.}, \textsc{Nowacki, A.~S.} and
  \textsc{Kattan, M.~W.} (2013). Strategies for handling missing data in
  electronic health record derived data. \textit{eGEMs}, \textbf{1}~(3).

\bibitem[{Williams and Hoel(2003)}]{williams2003modeling}
\textsc{Williams, B.~M.} and \textsc{Hoel, L.~A.} (2003). Modeling and
  forecasting vehicular traffic flow as a seasonal arima process: Theoretical
  basis and empirical results. \textit{Journal of transportation engineering},
  \textbf{129}~(6), 664--672.

\bibitem[{Wu \textit{et~al.}(2004)Wu, Ho and Lee}]{wu2004travel}
\textsc{Wu, C.-H.}, \textsc{Ho, J.-M.} and \textsc{Lee, D.-T.} (2004).
  Travel-time prediction with support vector regression. \textit{IEEE
  transactions on intelligent transportation systems}, \textbf{5}~(4),
  276--281.

\bibitem[{Wu and Tan(2016)}]{wu2016short}
\textsc{Wu, Y.} and \textsc{Tan, H.} (2016). Short-term traffic flow
  forecasting with spatial-temporal correlation in a hybrid deep learning
  framework. \textit{arXiv preprint arXiv:1612.01022}.

\bibitem[{Yin \textit{et~al.}(2002)Yin, Wong, Xu and Wong}]{yin2002urban}
\textsc{Yin, H.}, \textsc{Wong, S.}, \textsc{Xu, J.} and \textsc{Wong, C.}
  (2002). Urban traffic flow prediction using a fuzzy-neural approach.
  \textit{Transportation Research Part C: Emerging Technologies},
  \textbf{10}~(2), 85--98.

\bibitem[{Yu \textit{et~al.}(2017{\natexlab{a}})Yu, Wu, Wang, Wang and
  Ma}]{yu2017spatiotemporal}
\textsc{Yu, H.}, \textsc{Wu, Z.}, \textsc{Wang, S.}, \textsc{Wang, Y.} and
  \textsc{Ma, X.} (2017{\natexlab{a}}). Spatiotemporal recurrent convolutional
  networks for traffic prediction in transportation networks. \textit{Sensors},
  \textbf{17}~(7), 1501.

\bibitem[{Yu \textit{et~al.}(2017{\natexlab{b}})Yu, Li, Shahabi, Demiryurek and
  Liu}]{yu2017deep}
\textsc{Yu, R.}, \textsc{Li, Y.}, \textsc{Shahabi, C.}, \textsc{Demiryurek, U.}
  and \textsc{Liu, Y.} (2017{\natexlab{b}}). Deep learning: A generic approach
  for extreme condition traffic forecasting. In \textit{Proceedings of the 2017
  SIAM International Conference on Data Mining}, SIAM, pp. 777--785.

\bibitem[{Zarei \textit{et~al.}(2013)Zarei, Ghayour and
  Hashemi}]{zarei2013road}
\textsc{Zarei, N.}, \textsc{Ghayour, M.~A.} and \textsc{Hashemi, S.} (2013).
  Road traffic prediction using context-aware random forest based on volatility
  nature of traffic flows. In \textit{Asian Conference on Intelligent
  Information and Database Systems}, Springer, pp. 196--205.

\bibitem[{Zhao \textit{et~al.}(2017)Zhao, Chen, Wu, Chen and
  Liu}]{zhao2017lstm}
\textsc{Zhao, Z.}, \textsc{Chen, W.}, \textsc{Wu, X.}, \textsc{Chen, P.~C.} and
  \textsc{Liu, J.} (2017). Lstm network: a deep learning approach for
  short-term traffic forecast. \textit{IET Intelligent Transport Systems},
  \textbf{11}~(2), 68--75.

\end{thebibliography}


%
%
%
\end{document}